\documentclass[letterpaper]{article} 
\usepackage{aaai2027}  
\usepackage[hyphens]{url}  
\usepackage{graphicx} 
\usepackage{natbib}  
\usepackage{caption} 
\usepackage{algorithm}
\usepackage{algorithmic}

\usepackage{amssymb}
\usepackage{amsmath}
\usepackage{bm}
\usepackage{booktabs}
\usepackage{xcolor}
\usepackage{colortbl}
\usepackage{multirow}
\usepackage{pifont}
\usepackage{makecell}
\usepackage[most]{tcolorbox}
\usepackage{tikz}
\usepackage{amsthm}
\usepackage{graphicx}
\usepackage{subcaption}

\usepackage{newfloat}
\usepackage{listings}
\DeclareCaptionStyle{ruled}{labelfont=normalfont,labelsep=colon,strut=off}
\floatstyle{ruled}
\newfloat{listing}{tb}{lst}{}
\floatname{listing}{Listing}

\definecolor{green}{RGB}{0,140,0}
\definecolor{red}{RGB}{180,0,0}

\definecolor{findingbg}{RGB}{235,247,249}
\definecolor{findingframe}{RGB}{158,218,225}
\newtcolorbox{findingbox}{%
  enhanced, colback=findingbg, colframe=findingframe,
  leftrule=2pt, rightrule=0pt, toprule=0pt, bottomrule=0pt,
  arc=0pt, boxsep=2pt, left=8pt, right=8pt, top=6pt, bottom=6pt
}
\theoremstyle{plain}
\newtheorem{finding}{Question}

\makeatletter
\newcommand{\xdotfill}{\leavevmode\cleaders\hb@xt@ .6em{\hss.\hss}\hfill\kern\z@}
\makeatother

\newcommand{\listofapc}{%
  \clearpage
  \onecolumn
  \begin{center}
    \vspace*{0.5em}
    {\Huge\bfseries Appendix}\\[0.6em]
    {\large\scshape AgenticCANN: Automated Ascend C Operator Generation\\[0.2em] via Knowledge-Augmented Agentic Evolution}\\[0.8em]
    \rule{0.85\linewidth}{0.8pt}
  \end{center}
  \vspace{1.5em}
  \begin{center}
  \begin{minipage}{0.85\linewidth}
    \centerline{\Large\bfseries Table of Contents}
    \vspace{1.0em}
    \hrule height 0.6pt
    \vspace{1.0em}
    \normalsize
    \setlength{\parindent}{0pt}
    \setlength{\parskip}{0.35em}
    
    \textbf{A.}\enspace Related Work \dotfill \pageref{app:related_ext}
    \vspace{0.4em}
    
    \textbf{B.}\enspace Method Details \dotfill \pageref{app:method}
    
    \hspace{2.5em}\textbf{B.1}\enspace Algorithm Framework \dotfill \pageref{app:algorithm}
    
    \hspace{2.5em}\textbf{B.2}\enspace Knowledge Taxonomy \dotfill \pageref{app:knowledge_taxonomy}
    
    \hspace{2.5em}\textbf{B.3}\enspace Prompt Design and Agent Configuration \dotfill \pageref{app:prompts}
    
    \hspace{2.5em}\textbf{B.4}\enspace Hardware Constraints \dotfill \pageref{app:cann_arch}
    
    \hspace{2.5em}\textbf{B.5}\enspace Experimental Infrastructure \dotfill \pageref{app:infrastructure_details}
    \vspace{0.4em}
    
    \textbf{C.}\enspace Experimental Benchmarks \& Full Data Matrices \dotfill \pageref{app:experiments}
    
    \hspace{2.5em}\textbf{C.1}\enspace 54-Operator Stratified Dataset \dotfill \pageref{app:dataset_details}
    
    \hspace{2.5em}\textbf{C.2}\enspace Per-Operator Runtime Across Configurations \dotfill \pageref{app:full_runtimes}
    
    \hspace{2.5em}\textbf{C.3}\enspace Knowledge Injection Effect \& Replication \dotfill \pageref{app:knowledge_quantified}
    
    \hspace{2.5em}\textbf{C.4}\enspace Agent Interaction Traces \dotfill \pageref{app:traces_comparison}
    
    \hspace{2.5em}\textbf{C.5}\enspace Code Repair Case Study \dotfill \pageref{app:traces_casestudy}
    
    \hspace{2.5em}\textbf{C.6}\enspace 1B Pangu Profiling Details \dotfill \pageref{app:pangu_profiling}
    \vspace{0.4em}
    
    \textbf{D.}\enspace Discussion and Future Work \dotfill \pageref{app:discussion}
    
    \hspace{2.5em}\textbf{D.1}\enspace Key Findings and Implications \dotfill \pageref{app:findings}
    
    \hspace{2.5em}\textbf{D.2}\enspace Failure Analysis: Attention Operators \dotfill \pageref{app:failure_analysis}
    
    \hspace{2.5em}\textbf{D.3}\enspace Limitations \& Future Directions \dotfill \pageref{app:limitations}
    
    \vspace{1.0em}
    \hrule height 0.6pt
  \end{minipage}
  \end{center}
  \vspace{2.0em}
  \clearpage
}

\newcommand{\appsection}[1]{%
  \refstepcounter{section}%
  \setcounter{subsection}{0}%
  \addcontentsline{apc}{section}{\protect\numberline{\thesection}#1}%
  \section*{\thesection. #1}%
}
\newcommand{\appsubsection}[1]{%
  \refstepcounter{subsection}%
  \addcontentsline{apc}{subsection}{\protect\numberline{\thesubsection}#1}%
  \subsection*{\thesubsection. #1}%
}

\title{AgenticCANN: Automated Ascend C Operator Generation via Knowledge-Augmented Agentic Evolution}

\author{
    Junhao Qiu\textsuperscript{\rm 1},
    Zidong Wang\textsuperscript{\rm 1},
    Yansong Sun\textsuperscript{\rm 1},
    Zhitong Ma\textsuperscript{\rm 1},
    Ping Guo\textsuperscript{\rm 1,\rm 2},
    Qingfu Zhang\textsuperscript{\rm 1}\corresponding
}
\affiliations{
    \textsuperscript{\rm 1}Department of Computer Science, City University of Hong Kong\\
    \textsuperscript{\rm 2}Hong Kong Yuwin Technology Co., Ltd\\
    junhaoqiu2-c@my.cityu.edu.hk, qingfu.zhang@cityu.edu.hk
}

\begin{document}

\nocopyright
\maketitle

\begin{abstract}
Ascend C operator optimization is critical for NPU (Neural Processing Unit) inference performance but requires deep hardware expertise.
While large language models (LLMs) have shown promise in automated CUDA kernel generation, the fundamentally different programming model of Ascend C introduces unique challenges that remain unexplored.
In this paper, we propose \textbf{AgenticCANN}, a knowledge-augmented agentic evolution framework specifically tailored for automated Ascend C operator synthesis in low-corpus NPU environments.
To overcome the severe platform knowledge deficit on unfamiliar hardware, AgenticCANN incorporates a knowledge-orchestrated generation system that delivers structured, multi-level domain insights across the development lifecycle to resolve the upstream feasibility bottleneck.
Building on this foundation, it features a stage-adaptive agentic evolution strategy that dynamically aligns LLM interaction modes with specific generation and evolution phases, balancing high-exploration candidate discovery with high-convergence performance tuning.
Extensive experiments on Huawei Ascend 910B across six operators spanning five pattern categories demonstrate that our method achieves 90 to 100 percent feasibility on elementwise and normalization operators, 56\% on fusion operators, and up to 6.65$\times$ speedup on 1B Pangu model inference kernels. Further analysis reveals that knowledge injection monotonically improves feasibility from 57\% to 86\% on elementwise operators, demonstrating its general rather than operator-specific benefit.
\end{abstract}

\section{Introduction}\label{sec:intro}

\begin{figure*}
\centering
\includegraphics[width=1\textwidth]{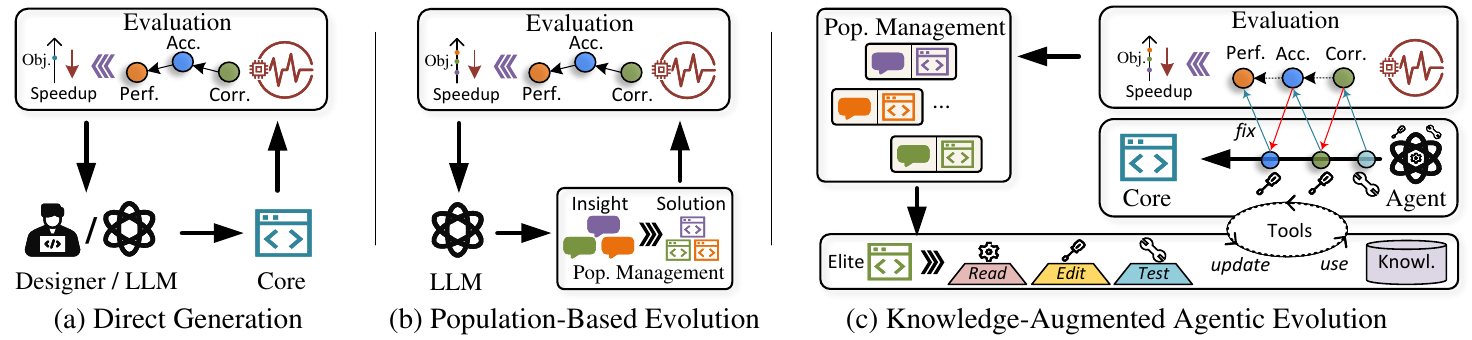}
\caption{Comparison of operator design paradigms: (a) direct operator design by human experts or single-pass LLM prompts, (b) population-based evolution leveraging collective heuristics, and (c) our proposed knowledge-augmented agentic evolution (AgenticCANN) empowered by active tool-use capabilities for low-corpus NPU environments.}
\label{fig:pipeline}
\end{figure*}

Large language models (LLMs)~\citep{touvron2023llama,guo2025deepseek} are driving transformative advances across software engineering, scientific discovery, and algorithm design, fueling unprecedented demand for efficient AI inference infrastructure.
As inference workloads diversify beyond NVIDIA GPUs, Huawei Ascend NPUs have emerged as a critical alternative, powered by the Ascend C programming model built on the CANN (Compute Architecture for Neural Networks) toolkit.
Ascend C offers a fundamentally different paradigm from CUDA: instead of fine-grained thread and warp control, it provides vector computation pipelines with unified buffer (UB) tiling, requiring programmers to explicitly manage data flow rather than parallel execution.
This architectural distinction creates both opportunities and challenges for automated code generation.

LLMs have demonstrated remarkable capabilities in automated CUDA kernel generation and optimization, with works such as KernelBench~\citep{ouyang2025kernelbench} and EvoEngineer~\citep{guo2025evoengineer} achieving substantial speedups by integrating tree search and population-based evolution with LLM-driven code mutation.
Foundational studies on code generation~\citep{chen2021evaluating,li2022competition} further establish the broad potential of LLMs for automated programming.
These approaches, however, share a critical assumption: the target platform (CUDA) is well-represented in the LLM's pre-training distribution, enabling the model to generate reasonable initial kernels and interpret compiler diagnostics.
On Ascend C, this assumption no longer holds.

Directly transferring these approaches to Ascend C thus faces a fundamental feasibility barrier.
While CUDA benefits from abundant open-source code in pre-training corpora, Ascend C represents a severe low-corpus domain with negligible public representation, causing existing frameworks to encounter an out-of-distribution generalization failure.
Without external knowledge injection, standard LLMs fail to generate even a single compilable solution for any operator beyond trivial elementwise patterns.
This shifts the primary challenge from the well-studied problem of kernel performance tuning to a more fundamental question: can LLMs achieve \textit{code synthesis feasibility} on unfamiliar hardware platforms under extreme data scarcity?

Prior work on LLM-based code evolution, such as EvoEngineer and Evolution of Heuristics~(EoH)~\citep{guo2025evoengineer,EOH}, exhibits two key limitations in such data-scarce scenarios.
First, as categorized in Figure~\ref{fig:pipeline}, while direct manual engineering (Figure~\ref{fig:pipeline}a) is unscalable, existing automated approaches (Figure~\ref{fig:pipeline}b) rely on closed-world population search where heuristics are evaluated strictly within task context and historical code.
While effective on data-rich platforms, this closed-world paradigm collapses when LLMs lack pre-trained domain representations, leaving raw compiler feedback uninterpretable without contextual domain knowledge.
Second, prior frameworks treat code evolution as a monolithic process where a uniform agent interaction archetype handles all search stages.
We observe that this monolithic strategy is suboptimal because initial solution discovery and subsequent kernel refinement require distinct cognitive capabilities, where a stage-adaptive agent design significantly improves both token efficiency and optimization convergence.

To address these limitations, we propose \textbf{AgenticCANN}, a novel framework featuring knowledge-augmented agentic evolution for low-corpus hardware platforms (Figure~\ref{fig:pipeline}c).
Our key insight is that structured domain knowledge orchestration transforms blind generation into guided search, while stage-adaptive agent scheduling optimizes resource allocation across the evolutionary lifecycle.

Our work makes three principal contributions:
\begin{itemize}
    \item \textbf{The AgenticCANN Framework for Low-Corpus NPU Environments:} We present AgenticCANN, an automated framework specifically designed for kernel code generation and evolution in low-corpus NPU scenarios. By bridging the gap between LLM capabilities and unfamiliar hardware specifications, AgenticCANN establishes a viable paradigm for automated Ascend C operator synthesis where pre-training data is severely scarce.

    \item \textbf{Knowledge-Orchestrated Generation and Stage-Adaptive Evolution:} We integrate knowledge-orchestrated generation with stage-adaptive agentic evolution into a unified pipeline. This design injects structured domain knowledge to overcome the upstream feasibility barrier, while dynamically scheduling high-exploration agents for initial seed discovery and high-convergence agents for downstream refinement, achieving +26.0\% improvement at 8.4\% of the token cost of monolithic approaches.

    \item \textbf{Empirical Validation on NPU Hardware:} Comprehensive validation across 54 stratified Ascend C operators spanning five pattern categories on Huawei Ascend 910B NPUs. AgenticCANN monotonically improves elementwise feasibility from 57\% to 86\%, enables reliable generation for production operators, and achieves up to 6.65$\times$ speedup on 1B Pangu model inference.
\end{itemize}

\section{Background and Problem Analysis}\label{sec:background}

\subsection{The Ascend C Programming Paradigm}
Unlike traditional GPU programming paradigms such as CUDA, which rely on fine-grained thread blocks and warp-level scheduling, the Huawei Ascend C programming model operates on an explicit dataflow architecture empowered by the CANN toolkit. Ascend C utilizes a vector compute pipeline combined with a multi-level memory hierarchy. 

The core programming abstraction revolves around UB tiling. 
Programmers are required to manually decompose input tensors into tile blocks that fit into the restricted UB capacity, manage ping-pong buffering queues (TQue, Task Queue) to overlap data transfer with vector computation, and insert explicit synchronization barriers between processing stages. 
This explicit hardware control eliminates implicit runtime overheads but imposes stringent syntactic and architectural constraints. 
Consequently, even minor misconfigurations in tile bounds, memory alignment, or queue synchronization cause catastrophic compilation failures or runtime memory illegal accesses.
Table~\ref{tab:platform_comp} summarizes the key architectural differences between CUDA and Ascend C that drive these challenges.

\begin{table}[t]
    \centering
    \small
    \begin{tabular}{l l l}
        \toprule
        Dimension & NVIDIA CUDA & Huawei Ascend C \\
        \midrule
        \makecell{Execution \\Model} & \makecell{Thread / Warp \\SIMT} & \makecell{Explicit TQue \\ Vector Pipeline} \\
        \midrule
        \makecell{Memory \\Hierarchy} & \makecell{Global / Shared \\ / Register} & \makecell{Global / UB \\ / L1} \\
        \midrule
        \makecell{Tiling \\Paradigm} & \makecell{Implicit, \\ thread-block scoped} & \makecell{Explicit \\ AlignUp Tiling} \\
        \midrule
        \makecell{Pre-training \\Corpus} & \makecell{Abundant \\(open-source)} & \makecell{Extremely Scarce \\(low-corpus)} \\
        \midrule
        \makecell{Primary \\Barrier} & \makecell{Downstream  \\latency tuning} & \makecell{Upstream compilation \\ feasibility} \\
        \bottomrule
    \end{tabular}
    \caption{Structural architectural comparison between CUDA (GPU) and Ascend C (NPU).}
    \label{tab:platform_comp}
\end{table}

\subsection{Problem Formulation}
We formulate the LLM-driven operator design problem as a constrained optimization task over a code candidate space $\mathcal{S}$. 
Given an operator specification $\mathcal{O}$ (comprising input/output tensor shapes, data types, and mathematical formulas), the goal is to discover a kernel implementation $x^* \in \mathcal{S}$ that satisfies correctness constraints while minimizing execution latency on the target hardware:
\begin{equation}\label{eq:formulation}
x^* = \arg\min_{x \in \mathcal{S}} T(x) \quad \text{subject to} \quad C(x) = 1,
\end{equation}
where $T(x)$ denotes the runtime execution latency measured on the NPU, and $C(x) \in \{0, 1\}$ is a binary indicator representing functional correctness and compilation success.

Existing automated code evolution frameworks (e.g., EvoEngineer, EoH) model this search process through a closed-world hypothesis. 
Specifically, the search state at iteration $t$ is updated purely using internal task context and historical candidate codes $\mathcal{H}_t$:
\begin{equation}
x_{t+1} \sim P_{\text{LLM}}(x \mid \mathcal{O}, \mathcal{H}_t, \text{Feedback}_t).
\end{equation}
While this closed-world loop succeeds on CUDA due to extensive open-source representation in $P_{\text{LLM}}$'s pre-training distribution, it completely breaks down in low-corpus domains like Ascend C. 
Due to extreme data scarcity, the pre-trained distribution lacks basic syntactic scaffolds and API representations for Ascend C. 
Under such conditions, the probability of sampling a functionally valid candidate without external domain knowledge approaches zero for non-trivial operators:
\begin{equation}
P_{\text{LLM}}(C(x) = 1 \mid \mathcal{O}) \approx 0, \quad \forall x \notin \text{Elementwise}.
\end{equation}
As a result, raw compiler error logs offer uninterpretable feedback to the LLM, trapping closed-world evolutionary search in a perpetual compilation failure loop. 
This establishes that the primary challenge in low-corpus hardware synthesis is an upstream \textit{feasibility bottleneck} rather than downstream performance tuning.

\subsection{Exploration vs. Convergence Trade-off}
To overcome the feasibility barrier and evolve valid kernels, LLMs can be deployed via various interaction archetypes, ranging from simple fix loops to autonomous tool-using agents. 
However, applying a uniform (monolithic) agent mode across the entire evolutionary lifecycle introduces an inherent trade-off between exploration and convergence.

We characterize three representative agent interaction archetypes along a capability spectrum:
\begin{enumerate}
    \item \textbf{Single-Pass / Lightweight Fix Loops:} Low token cost and high structural stability during iterative refinement, but virtually incapable of synthesizing viable initial code structures from scratch in data-scarce search spaces.
    \item \textbf{ReAct CodeAgents:} Intermediate capability using structured reasoning-action cycles; capable of repairing minor bugs but susceptible to local optima.
    \item \textbf{Full Tool-Using Agents (e.g., Smolagents):} Unrestricted environment tool usage and execution feedback. They possess powerful exploratory capabilities to establish valid initial code structures, but introduce severe structural drift
\end{enumerate}

\section{The AgenticCANN Framework}\label{sec:method}

To address the severe feasibility barrier and agency mismatch inherent in low-corpus NPU programming, we propose \textbf{AgenticCANN}, an automated kernel generation and evolutionary optimization framework. The framework directly tackles the upstream feasibility bottleneck (formalized in Section~2): without structured external knowledge, $P_{\text{LLM}}(C(x){=}1 \mid \mathcal{O}) \approx 0$ for non-trivial operators. AgenticCANN breaks this deadlock through two complementary mechanisms. First, \textit{knowledge-orchestrated generation} (Section~\ref{subsec:knowledge}) proactively delivers multi-level domain insights across evolutionary phases, transforming blind generation into guided search. Second, \textit{stage-adaptive agentic evolution} (Section~\ref{subsec:agency}) dynamically matches agent capabilities to search phases: high-exploration modes for initial seed discovery where the feasibility barrier is highest, followed by high-convergence modes for cost-effective refinement.

\begin{figure*}
\centering
\includegraphics[width=1\textwidth]{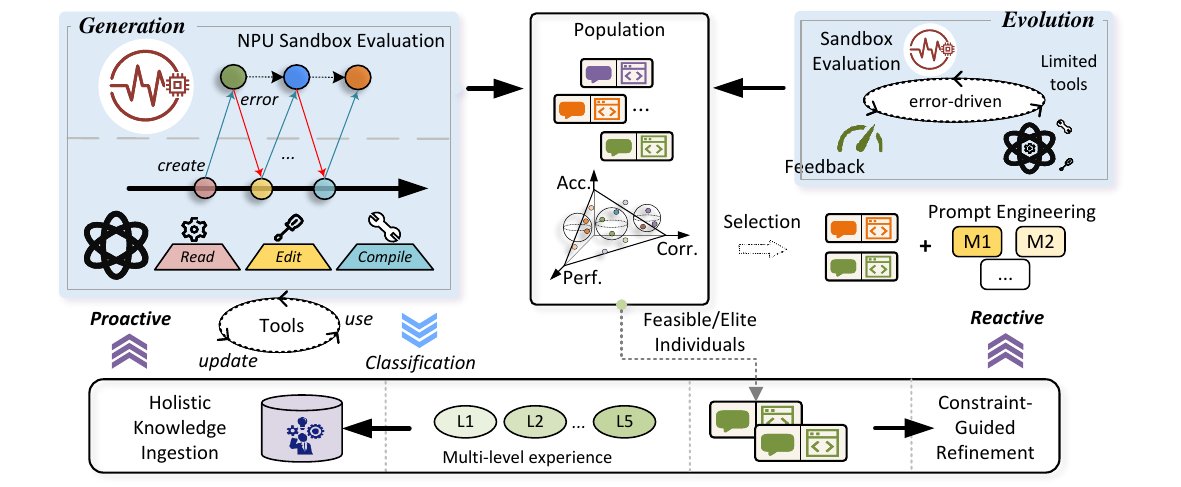}
\caption{Framework Overview. AgenticCANN employs a dual-phase architecture: knowledge-orchestrated generation at $g=1$ injects full domain taxonomy to overcome the feasibility barrier, while stage-adaptive evolution for $g>1$ dynamically schedules agent modes for efficient convergence.}
\label{fig:framework}
\end{figure*}

\begin{table}[ht]
    \centering
    \small
    \begin{tabular}{c c c c}
        \toprule
        Level & Category & Scope \& Content & Cognitive Function \\
        \midrule
        L0 & \makecell{Prog. \\Model} & \makecell{Pipelines, TQue, \\ UB memory} & \makecell{Establishes hardware \\ mental model} \\
        \midrule
        L1 & \makecell{Pattern \\Guide} & \makecell{Mapping rules, \\ tensor layout} & \makecell{Bridges math \\ semantics to code} \\
        \midrule
        L2 & \makecell{HW \\Constraint} & \makecell{UB limits, \\ alignment, buffers} & \makecell{Enforces execution \\ invariants} \\
        \midrule
        L3 & \makecell{Tiling \\Strategy} & \makecell{Tiling formulas, \\ tail handling} & \makecell{Directs memory \\ access throughput} \\
        \midrule
        L4 & \makecell{API \\Reference} & \makecell{Vector API \\ signatures} & \makecell{Ensures syntactic \\ validity} \\
        \midrule
        L5 & Exemplars & \makecell{Verified operator \\ kernel code} & \makecell{Provides structural \\ scaffolds} \\
        \bottomrule
    \end{tabular}
    \caption{Multi-level knowledge taxonomy for hardware programming.}
    \label{tab:knowledge_taxonomy}
\end{table}

Algorithm~\ref{alg:framework_app} formalizes the complete workflow. At each generation, the framework infers the compute pattern $\pi$, assembles a phase-differentiated knowledge context $\mathcal{K}_g$, and deploys a stage-appropriate interaction mode $\text{mode}_g$ (lines~3--5). Candidates undergo a sandboxed three-stage pipeline (compilation, numerical verification, hardware profiling on physical NPUs), with errors triggering targeted correction routines grounded in specialized hardware constraints (lines~11--14).

Architecturally, AgenticCANN consists of three tightly coupled functional modules (Figure~\ref{fig:framework}):
\begin{enumerate}
    \item \textbf{Knowledge Layer:} Manages a multi-level domain taxonomy and executes pattern-based routing with phase-differentiated assembly, supplying non-pretrained domain rules to guide code generation.
    \item \textbf{Generation and Evolution Layer:} Hosts a spectrum of stage-adaptive interaction modes, spanning single-pass synthesis, feedback-driven error correction, and full tool-using agents, which are dynamically scheduled across search stages.
    \item \textbf{Evaluation Layer:} Orchestrates a sandboxed compile-verify-measure workflow using physical device pooling, streaming diagnostic traces back to the Knowledge Layer to direct structural kernel repairs.
\end{enumerate}

\subsection{Knowledge-Orchestrated Generation}\label{subsec:knowledge}

A fundamental limitation of reactive feedback in low-corpus domains is that the LLM cannot interpret hardware-specific compiler diagnostics without prior domain grounding. To establish effective synthesis, hardware constraints must be injected proactively before code generation..

\subsubsection{Multi-Level Knowledge Taxonomy}

Raw technical documentation for specialized hardware is typically voluminous and uncurated. 
Directly appending entire user guides into prompts causes severe context window dilution, causing the model to miss critical hardware constraints. 
To structure this knowledge effectively, we formulate a six-level cognitive taxonomy (Table~\ref{tab:knowledge_taxonomy}) spanning hardware mental models (L0) through operational semantics (L1), invariant constraints (L2), tiling strategies (L3), API specifications (L4), and verified structural exemplars (L5).

\subsubsection{Phase-Differentiated Assembly}

Supplying all six taxonomy levels across all evolutionary steps introduces redundant context that impairs model focus. We resolve this through phase-differentiated assembly (detailed in Appendix~\ref{tab:phase_assembly}): each evolutionary phase receives only the knowledge levels relevant to its task. Initialization injects full knowledge (L0--L5) to construct the initial mental model. Mutation restricts context to compact L2--L4 guardrails to prevent structural drift. Compilation repair focuses on structural examples and hardware constraints to fix queue and API errors. Correctness repair combines pattern guides with complete examples to resolve numerical discrepancies. Performance tuning isolates tiling strategies to guide memory bandwidth optimization. By filtering out non-essential documentation, this mechanism optimizes context precision and minimizes prompt overhead. Prompt length drops from roughly 800 tokens at initialization to under 100 tokens at performance tuning, an 8$\times$ reduction in per-generation cost.

\subsubsection{Pattern-Based Routing}

Optimization techniques depend on the specific mathematical and memory characteristics of an operator. 
For instance, tiling strategies for reduction operations differ fundamentally from those used in elementwise transformations. 
AgenticCANN maintains a pattern-based routing system that classifies target operators into canonical compute categories, including elementwise, reduction, softmax, broadcast, normalization, matrix multiplication, convolution, attention, index, pooling, resize, and fusion patterns.

Each compute pattern maps to dedicated knowledge assets across all six taxonomy levels. 
Given a target specification $\mathcal{O}$, the framework infers its compute pattern $\pi$, retrieves the corresponding asset files, and constructs a context tailored to the active search phase. 
This decoupling ensures that supporting new hardware operators requires only adding pattern-specific knowledge assets, leaving the core routing and evolutionary algorithms unchanged.

\subsection{Stage-Adaptive Evolution}\label{subsec:agency}

Iterative program evolution requires balancing exploration and convergence. 
Initial candidate synthesis demands high exploratory freedom to overcome the feasibility barrier and discover valid code seeds. 
Conversely, subsequent performance optimization requires structural convergence, where targeted mutations refine execution efficiency without destroying functional correctness.

We formalize three interaction archetypes along a capability spectrum:
\begin{itemize}
    \item \textbf{Single-Pass Mode:} Synthesizes complete source code in a single turn without feedback. It preserves structural stability and minimizes token overhead, but exhibits low exploration capabilities when resolving initial syntax errors in data-scarce settings.
    \item \textbf{Fix-Loop Mode:} Integrates closed-loop feedback into Single-Pass generation. Upon compilation or runtime failure, execution logs and phase-specific knowledge contexts are provided to repair code iteratively.
    \item \textbf{Tool-Agent Mode:} Equips the model with interactive tools for file inspection (\texttt{read}), localized modification (\texttt{edit}), and compiler execution (\texttt{compile}). This mode provides strong exploration capabilities, but can introduce destructive structural mutations that disrupt evolutionary convergence.
\end{itemize}

\begin{table}[t]
    \centering
    \footnotesize 
    \setlength{\tabcolsep}{5.5pt} 
    \begin{tabular}{l l l c c}
        \toprule
        Mode & Tools & Feedback Source & Expl. & Conv. \\
        \midrule
        Single-Pass & None & None & Low & High \\
        Fix-Loop & None & Logs + Context & Med & Med \\
        Tool-Agent & R / E / C & Execution trace & High & Low \\
        \bottomrule
    \end{tabular}
    \caption{Spectrum of agent interaction modes (Tools: R -- Read, E -- Edit, C -- Compile; Expl.: Exploration capability; Conv.: Convergence stability). High exploration capability trades off against structural convergence stability.}
    \label{tab:generation_modes}
\end{table}

AgenticCANN resolves this trade-off using a stage-adaptive scheduling policy: during initialization ($g=1$), high-exploration modes (Tool-Agent or Fix-Loop) establish valid candidate seeds; in subsequent generations ($g>1$), the framework transitions to high-convergence modes (Single-Pass or lightweight Fix-Loop) for structure-preserving optimization. 
This dynamic scheduling aligns agent capabilities with phase-specific search requirements. On GELU, stage-adaptive scheduling achieves +26.0\% improvement with 350K tokens, versus +0.04\% for a monolithic Tool-Agent consuming 4.18M tokens (Table~\ref{tab:gelu}).

\section{Experiments}\label{sec:exp}

We evaluate AgenticCANN on three dimensions: knowledge orchestration (Section~4.2), stage-adaptive evolution (Section~4.3), and production deployment (Section~4.4).

\subsection{Experimental Setup}

All benchmarking experiments are executed on a dedicated cluster containing eight Huawei Ascend 910B1 NPUs (61\,GB HBM, High Bandwidth Memory, per card) running CANN 8.1.RC1, with two physical NPU devices allocated per experimental job. 
Large Language Model capabilities are accessed via API using DeepSeek-V4-Flash for the initial six-operator in-depth study and DeepSeek-V4-Pro for the large-scale 54-operator validation. 
Sampling parameters are set to temperature $T=0.8$, top\_p $=0.95$, and maximum generation length of $8192$ tokens. 
Full system prompts and configuration specifications are detailed in Appendix~\ref{app:knowledge_taxonomy}.

We conduct two complementary suites of evaluations:
\begin{enumerate}
    \item \textbf{In-Depth Representative Operator Benchmark:} An intensive diagnostic analysis across six canonical operators representing distinct execution patterns, including GELU (elementwise), LayerNorm (normalization), RMSNorm (normalization), Softmax (reduction), ResidualAdd+RMSNorm (fusion), and add\_bias\_broadcast (broadcast).
    \item \textbf{Large-Scale Stratified Benchmark:} A broad generalization evaluation spanning 54 Ascend C operators grouped by computational topology (detailed in Appendix~\ref{app:dataset_details}).
\end{enumerate}

Primary evaluation metrics include: \textit{Feasibility Rate} (the percentage of synthesis trials producing fully compilable and numerically verified solutions), \textit{Speedup} (the ratio of baseline execution latency to optimized kernel latency, averaged over 10 execution runs), and \textit{Token Overhead} (cumulative LLM API consumption). 
We benchmark AgenticCANN against standard closed-loop evolutionary baselines, specifically Evolution of Heuristics (EoH)~\citep{EOH}. 
Evolutionary parameters are set to a population size of 5 candidates, running for 5 generations with 2 parallel sampling workers, 3 numerical correctness verification passes, and a 3600-second execution sandbox timeout (Appendix~\ref{app:infrastructure_details}).

\subsection{Evaluating Knowledge Orchestration}\label{subsec:exp_knowledge}

\subsubsection{Overall Feasibility across Operator Patterns}

Table~\ref{tab:feasibility} presents performance metrics across six representative operators. 
The empirical results confirm that the platform knowledge deficit creates an absolute feasibility barrier. 
Without proactive knowledge orchestration, standard closed-world generation fails completely, yielding a 0\% feasibility rate across all non-elementwise operators. 
The LLM cannot synthesize a single valid normalization, reduction, or fused operator kernel using pretrained knowledge alone.
\begin{table}[t]
    \centering
    \footnotesize 
    \setlength{\tabcolsep}{4.5pt} 
    \begin{tabular}{l c l c c}
        \toprule
        Operator & Type & Mode & Feasibility & Speedup \\
        \midrule
        GELU & I & Fix-Loop & \textbf{100\%} & \textbf{2.71$\times$} \\
        GELU & I & Tool-Agent & \textbf{100\%} & 1.63$\times$ \\
        GELU & I & EoH & 83\% & 0.99$\times$ \\
        \midrule
        LayerNorm & II & Fix-Loop & \textbf{90\%} & \textbf{2.71$\times$} \\
        RMSNorm & II & Fix-Loop & \textbf{100\%} & \textbf{1.41$\times$} \\
        \midrule
        Softmax & III & Fix-Loop & \textbf{100\%} & \textbf{0.98$\times$} \\
        \midrule
        Fused-AddNorm & IV & Fix-Loop & \textbf{56\%} & \textbf{1.10$\times$} \\
        \midrule
        Add-Broadcast & V & All Modes & 0\% & --- \\
        \bottomrule
    \end{tabular}
    \caption{Overall feasibility and speedup across canonical operator patterns (Types: I -- Elementwise, II -- Normalization, III -- Reduction, IV -- Fusion, V -- Broadcast).}
    \label{tab:feasibility}
\end{table}

When structured knowledge orchestration is applied, elementwise and normalization operators achieve high feasibility rates between 90\% and 100\%, accompanied by speedup gains up to 2.71$\times$. 
Complex operator fusions reach a 56\% feasibility threshold, whereas broadcast operations remain at 0\%, highlighting current boundaries in handling dynamic index alignment.

\subsubsection{Knowledge Level Ablation Analysis}

To isolate the marginal contribution of each knowledge level, Table~\ref{tab:knowledge} reports ablation results on LayerNorm, a representative Level~3 operator requiring explicit pipeline buffer allocation.

\begin{table}[t]
    \centering
    \small
    \begin{tabular}{l c c}
        \toprule
        Knowledge Configuration & Feasibility & Speedup \\
        \midrule
        Closed-World (No domain know.) & 0\% & --- \\
        \midrule
        + L0 (Programming Model) & 40\% & 1.12$\times$ \\
        \midrule
        + L0 + L2 (Hardware Constraints) & 60\% & 1.85$\times$ \\
        \midrule
        + L0 + L2 + L5 (Curated Example) & 80\% & 2.41$\times$ \\
        \midrule
        Full (Phase-Differentiated) & \textbf{90\%} & \textbf{2.71$\times$} \\
        \bottomrule
    \end{tabular}
    \caption{Knowledge taxonomy ablation on LayerNorm. Invariant constraints (L2) yield the largest single marginal improvement.}
    \label{tab:knowledge}
\end{table}

\begin{figure}[t]
    \centering
    \includegraphics[width=\columnwidth]{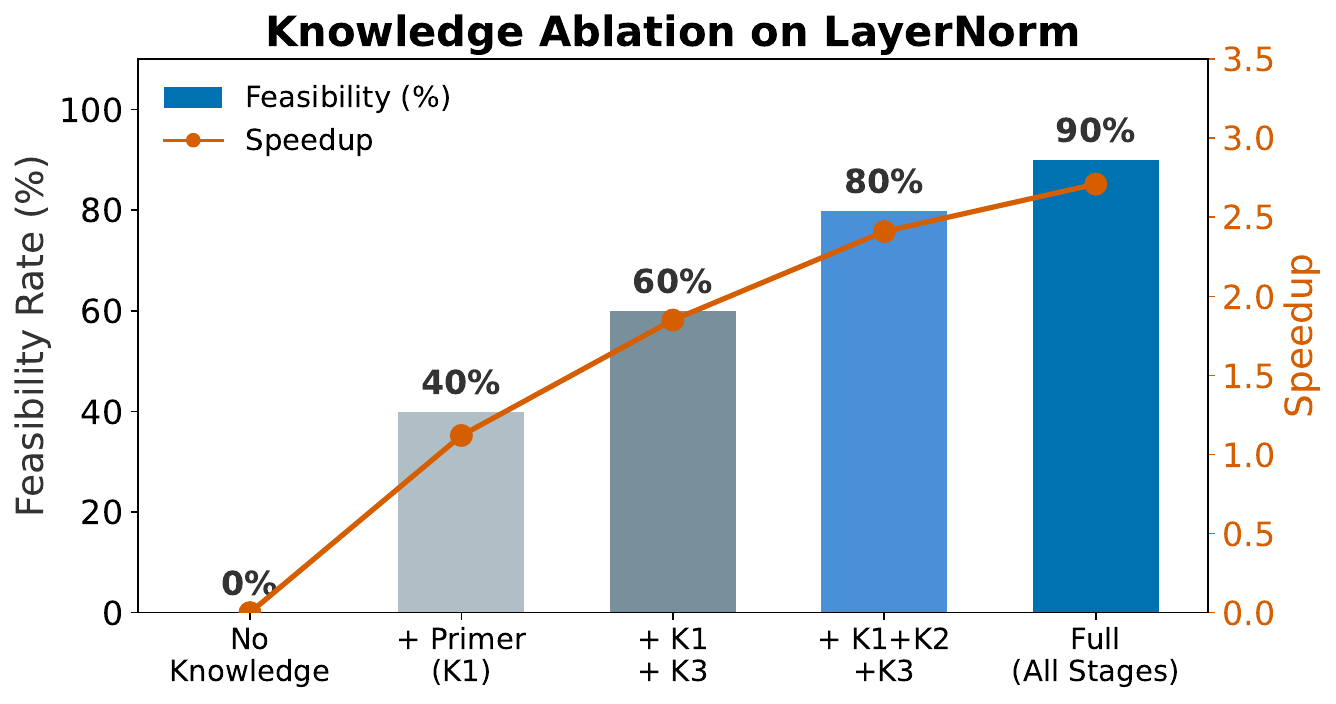}
    \caption{Knowledge ablation on LayerNorm. Bars show feasibility rate; the line shows speedup. Each additional knowledge type improves both metrics. Full knowledge across all stages achieves 90\% feasibility and 2.71$\times$ speedup.}
    \label{fig:knowledge_ablation}
\end{figure}

The progression from 0\% to 90\% feasibility illustrates how structured domain context unlocks generation capacity. 
Injecting L0 (Programming Model) establishes basic hardware abstractions, raising feasibility to 40\%. 
Adding L2 (Hardware Constraints) provides the single largest marginal gain, increasing feasibility to 60\% and proving that explicit memory alignment rules are critical for low-level kernel compilation. 
Incorporating L5 (Curated Examples) increases feasibility to 80\% by supplying concrete implementation backbones. 
Finally, deploying the full phase-differentiated assembly achieves a peak 90\% feasibility rate and a 2.71$\times$ execution speedup. Figure~\ref{fig:knowledge_ablation} visualizes this monotonic progression.

\subsubsection{Large-Scale Benchmark Generalization}

To evaluate whether knowledge orchestration generalizes across broad operator suites, we conduct a stratified evaluation across 54 operator kernels under three distinct knowledge regimes: \textit{Free} (closed-world generation without knowledge), \textit{Insight} (supplying only Level 5 curated examples), and \textit{Full} (applying full taxonomy with phase-differentiated assembly).

\begin{table}[t]
    \centering
    \small
    \begin{tabular}{l c c}
        \toprule
        Configuration & Valid Solutions & Avg. Latency \\
        \midrule
        Free (No domain know.) & 8/14 (57\%) & 741.3\,$\mu$s \\
        \midrule
        Insight (+ L5 Exemplars) & 10/14 (71\%) & 721.6\,$\mu$s \\
        \midrule
        Full (Phase-Differentiated) & \textbf{12/14 (86\%)} & \textbf{711.0\,$\mu$s} \\
        \bottomrule
    \end{tabular}
    \caption{Stratified benchmark across 14 elementwise operator kernels.}
    \label{tab:knowledge_scale}
\end{table}

\begin{figure}[ht]
    \centering
    \includegraphics[width=\columnwidth]{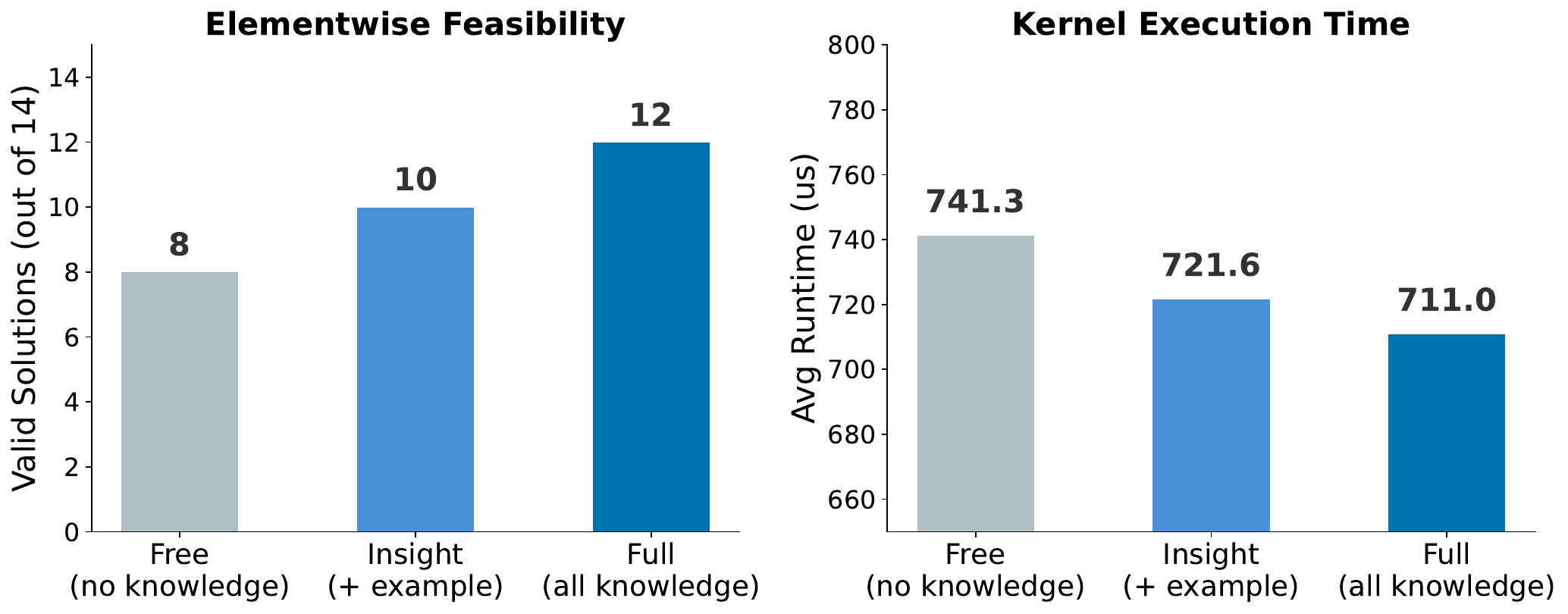}
    \caption{Large-scale knowledge injection validation across 14 elementwise operators. Left: valid solution count monotonically increases from 8/14 to 12/14. Right: average kernel execution time decreases from 741.3$\mu$s to 711.0$\mu$s.}
    \label{fig:knowledge_scale}
\end{figure}

\begin{figure}[t]
    \centering
    \includegraphics[width=\columnwidth]{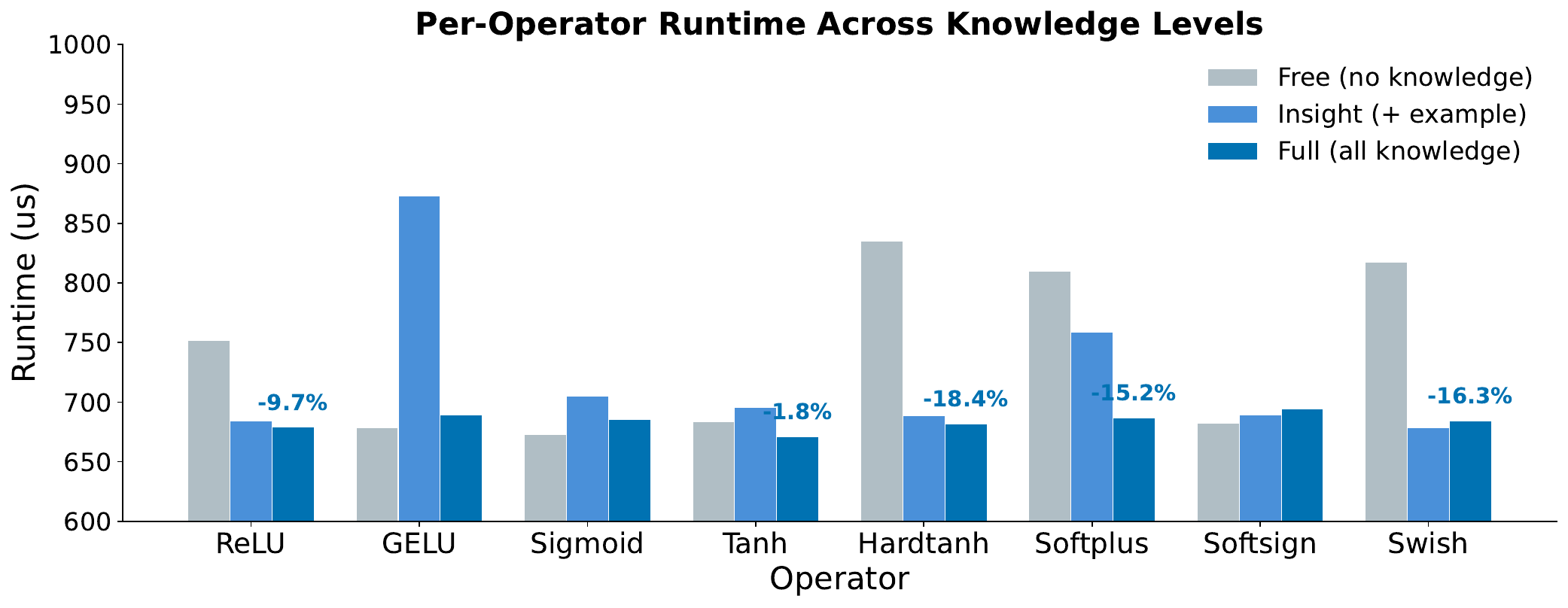}
    \caption{Per-operator runtime comparison across three knowledge levels. Knowledge injection reduces execution time for most operators, with hardtanh showing 18.4\% improvement and softplus showing 15.2\% improvement.}
    \label{fig:per_op_runtime}
\end{figure}

As summarized in Table~\ref{tab:knowledge_scale}, expanding domain context monotonically improves functional validity from 57\% (Free) to 71\% (Insight), reaching 86\% under Full knowledge orchestration. 
Simultaneously, mean kernel execution latency decreases from 741.3\,$\mu$s to 711.0\,$\mu$s. 
Individual operator evaluation demonstrates consistent performance enhancements under Full knowledge: Hardtanh latency drops from 835.3\,$\mu$s to 681.5\,$\mu$s (a 18.4\% reduction), Softplus decreases from 809.8\,$\mu$s to 686.9\,$\mu$s (a 15.2\% reduction), Swish declines from 817.4\,$\mu$s to 678.3\,$\mu$s (a 17.0\% reduction), and ReLU improves from 751.7\,$\mu$s to 678.8\,$\mu$s (a 9.7\% reduction). Figure~\ref{fig:knowledge_scale} visualizes the aggregate trend across all 14 operators, and Figure~\ref{fig:per_op_runtime} provides the per-operator breakdown.

\subsection{Validating Stage-Adaptive Evolution}\label{subsec:exp_agency}

To validate the hypothesis that agent agency must be dynamically matched to the search phase, Table~\ref{tab:gelu} benchmarks interaction archetypes on the GELU operator over a multi-generation evolutionary run.

\begin{table}[ht]
    \centering
    \footnotesize 
    \setlength{\tabcolsep}{6pt} 
    \begin{tabular}{l c c c}
        \toprule
        Metric & Fix-Loop & Tool-Agent & EoH \\
        \midrule
        Feasibility Rate & \textbf{100\%} & \textbf{100\%} & 83\% \\
        Best Latency ($\text{ms}$) & 36.12 & \textbf{30.37} & 48.84 \\
        Evo. Gain ($\%$) & \textbf{+26.0} & +0.04 & 0.00 \\
        Token Cost & 350K & 4.18M & \textbf{242K} \\
        \bottomrule
    \end{tabular}
    \caption{Performance comparison across interaction archetypes during kernel evolution (Evo. Gain: Evolutionary Optimization Gain over initial candidates; Token Cost: Cumulative LLM token overhead).}
    \label{tab:gelu}
\end{table}

The empirical findings highlight a key trade-off between agent exploration and evolutionary convergence. 
Tool-Agent mode synthesizes a high-performing single kernel candidate (30.37\,ms), illustrating its strong initial exploration capability. 
However, across subsequent generations, Tool-Agent yields a negligible +0.04\% performance improvement despite consuming 4.18M tokens. 
Unconstrained tool interactions introduce broad structural alterations that disrupt incremental performance optimization. 
Conversely, Fix-Loop mode maintains high structural stability during evolution, achieving a +26.0\% performance gain over generations while consuming only 350K tokens, representing a fraction of Tool-Agent's token cost. 
These results confirm that high-agency modes should be restricted to initial seed drafting, whereas structured, low-agency modes are necessary for reliable evolutionary convergence.

\subsection{Production Deployment and Effectiveness}\label{subsec:exp_practical}

\subsubsection{Error-Driven Kernel Refinement}

The feedback refinement loop demonstrates that targeted knowledge injection is vital during candidate repair. 
Table~\ref{tab:perffix} tracks performance optimization during execution error correction iterations.

\begin{table}[ht]
    \centering
    \small
    \begin{tabular}{l c c c}
        \toprule
        Pattern & Initial Speedup & Speedup & Iterations \\
        \midrule
        LayerNorm & 0.39$\times$ & \textbf{2.68$\times$} & \textbf{1} \\
        \midrule
        RMSNorm & $<$ 1.00$\times$ & \textbf{1.41$\times$} & 4 \\
        \midrule
        Softmax & $<$ 1.00$\times$ & \textbf{0.98$\times$} & 4 \\
        \midrule
        Fused-AddNorm & $<$ 1.00$\times$ & \textbf{1.10$\times$} & \textbf{1} \\
        \bottomrule
    \end{tabular}
    \caption{Performance progression during error-driven repair iterations.}
    \label{tab:perffix}
\end{table}

\begin{table}[ht]
    \centering
    \small
    \begin{tabular}{l c c c}
        \toprule
        Tensor Shape & PyTorch & AgenticCANN & Speedup \\
        \midrule
        (1, 1, 1536) & 0.118\,ms & \textbf{0.018\,ms} & \textbf{6.65$\times$} \\
        \midrule
        (1, 26, 1536) & 0.108\,ms & \textbf{0.017\,ms} & \textbf{6.19$\times$} \\
        \bottomrule
    \end{tabular}
    \caption{Execution latency benchmarks for RMSNorm on 1B Pangu model shapes.}
    \label{tab:pangu}
\end{table}
Providing Level 3 tiling heuristics during repair allows the LayerNorm kernel to improve from an unoptimized 0.39$\times$ baseline to a 2.68$\times$ speedup in a single iteration. 
RMSNorm achieves a 1.41$\times$ speedup after 4 refinement steps. 
Softmax achieves a 0.98$\times$ ratio, indicating that vendor compiler passes already apply aggressive optimizations to standard reduction topologies, limiting further headroom.

\subsubsection{End-to-End Production Deployment on 1B Pangu}

\begin{figure}[ht]
    \centering
    \includegraphics[width=\columnwidth]{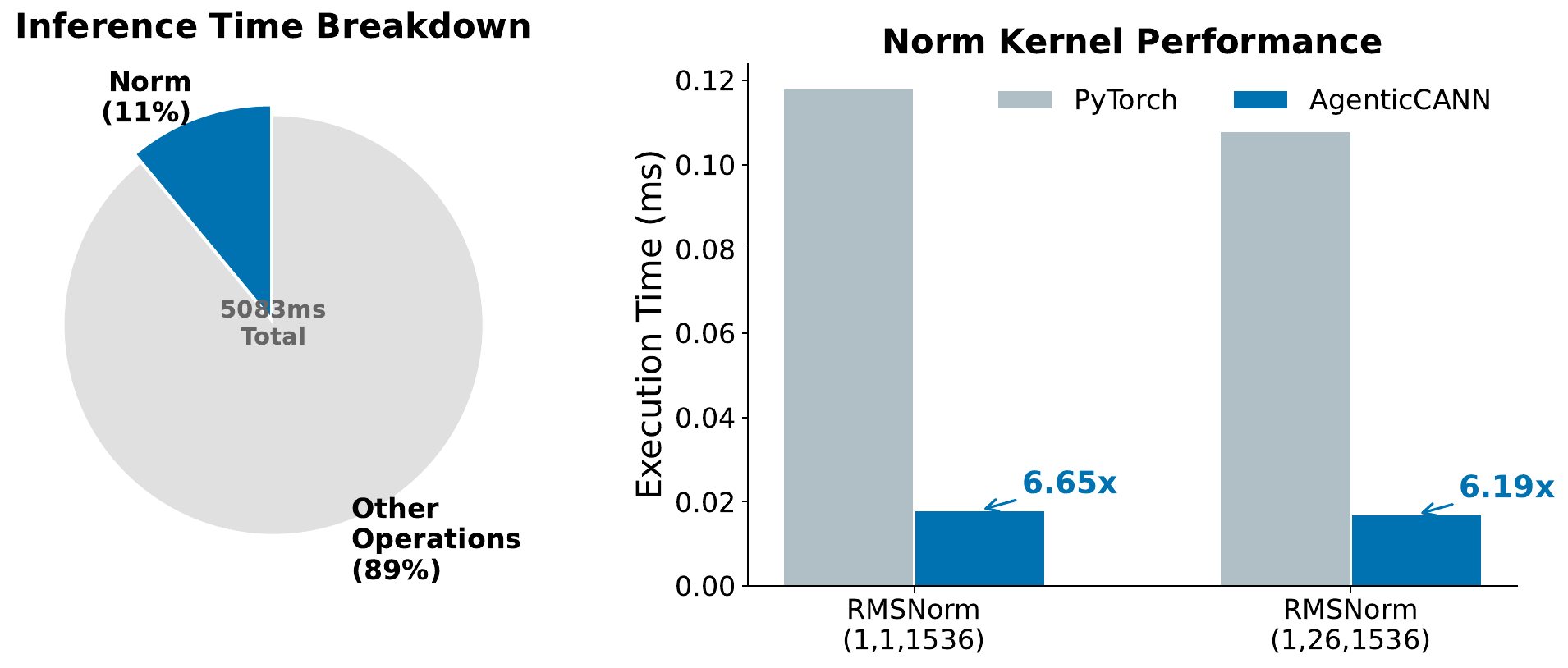}
    \caption{1B Pangu inference profiling. Left: time breakdown showing Norm accounts for 11\% of total inference time (5083ms). Right: Norm kernel performance showing 6.19 to 6.65$\times$ speedup over PyTorch at inference shapes.}
    \label{fig:pangu_combined}
\end{figure}

To evaluate real-world impact, optimized kernels were integrated directly into the inference pipeline of the 1B parameter Pangu language model executed on physical Ascend 910B NPUs~\cite{chen2025pangu}..
Profiling analysis reveals that normalization kernels account for approximately 11\% of total model end-to-end execution latency, consuming 1102\,ms across 3392 layer invocations within a full generation pass. 
Table~\ref{tab:pangu} details single-kernel execution benchmarks comparing standard PyTorch implementations against optimized Ascend C kernels across production inference tensor dimensions.
The synthesized RMSNorm kernel achieves up to a 6.65$\times$ execution speedup over standard PyTorch operators at production inference shapes. 
Replacing all 53 normalization layers across the transformer backbone yields an estimated 8--10\% reduction in end-to-end model inference latency. This demonstrates that LLM-generated kernels, produced without manual hardware tuning, can deliver practical speedups on production-grade models. Figure~\ref{fig:pangu_combined} provides the full profiling breakdown.

\section{Conclusion}\label{sec:conclusion}

This paper presented AgenticCANN, an automated code generation and evolutionary optimization framework designed for specialized NPU computing architectures operating under low-corpus data regimes. 
To address the foundational platform knowledge deficit, we introduced a structured six-level domain taxonomy coupled with a phase-differentiated assembly mechanism. 
This proactive knowledge orchestration elevates kernel synthesis feasibility for complex normalization operators from 0\% to 90\% without requiring parametric model adaptation. 
To resolve the operational tension between exploratory agency and evolutionary convergence, we introduced a stage-adaptive scheduling policy that pairs high-exploration interaction modes during candidate initialization with high-convergence modes during multi-generation refinement. 
Extensive benchmarking across 54 operators on Huawei Ascend 910B NPUs validates substantial gains in candidate feasibility, along with execution speedups up to 6.65$\times$ when integrated into production LLM inference pipelines. 
Future work will focus on automating knowledge extraction directly from diagnostic compilation logs and extending the framework to orchestrate end-to-end attention and convolution computational topologies.

\bibliography{ref}


\listofapc
\appendix
\setcounter{page}{1}

\appsection{Related Work}\label{app:related_ext}

\medskip
\noindent\textbf{LLM-based Kernel Optimization.}
Large language models have demonstrated remarkable capabilities in general software engineering~\citep{chen2021evaluating,li2022competition,roziere2023code}, functional code synthesis~\citep{roziere2023code}, and mathematical algorithm discovery~\citep{funsearch,novikov2025alphaevolve}. 
Building upon these capabilities, recent literature has extended LLM applications toward hardware-level program synthesis and kernel optimization. 
Benchmark suites such as KernelBench~\citep{ouyang2025kernelbench} and TritonBench~\citep{li2025tritonbench} systematically evaluate LLMs in generating high-performance compute kernels across heterogeneous acceleration platforms. 
To go beyond static generation, automated optimization frameworks employ iterative refinement loops. 
For instance, EvoEngineer~\citep{guo2025evoengineer} introduces population-based tree search to evolve CUDA kernels, while related approaches explore multi-objective trade-offs and automated heuristic design for scheduling problems~\citep{qiu2026llm,qiu2026e2oc,wang2025llm}. 
However, a pervasive limitation of these methodologies is their underlying assumption of a data-rich closed-world paradigm. 
They rely heavily on the vast pre-training knowledge of CUDA present in public repositories, allowing code evolution to succeed through self-contained heuristic mutations. 
When transferred to low-corpus NPU architectures like Huawei Ascend C, where training data is virtually nonexistent, this closed-world assumption breaks down completely, resulting in a severe feasibility bottleneck for complex operators.

\medskip
\noindent\textbf{Overcoming Data Scarcity via Knowledge.}
To mitigate domain knowledge gaps without full fine-tuning, retrieval-augmented generation (RAG)~\citep{lewis2020retrieval} and in-context learning tactics~\citep{liu2023pre} have been widely adopted. 
In code intelligence tasks, injecting structural scaffolds, API documentation, and curated code demonstrations significantly reduces syntax hallucinations and improves initial compilation success~\citep{madaan2023self}. 
In domain-specific environments, structured knowledge management has also proven effective for complex constraint satisfaction and system modeling~\citep{gao2023retrieval}. 
Nevertheless, existing knowledge injection techniques typically treat external domain information as flat, static text appended uniformly to prompt contexts. 
They lack hardware-aware taxonomy systems that categorize domain rules according to NPU hardware execution constraints, such as unified buffer (UB) tiling, vector pipeline queues, and synchronization barriers. 
Furthermore, static injection approaches fail to adapt knowledge delivery to different evolutionary stages, leading to severe prompt inflation and context dilution.

\medskip
\noindent\textbf{Agentic Workflows and Stage Evolution.}
The evolution of LLM interaction paradigms has shifted from single-pass prompting to autonomous agentic workflows~\citep{yao2022react,wang2024survey}. 
In automated program synthesis, agentic loops incorporate execution feedback, unit test results, and compiler logs to iteratively repair buggy code~\citep{shinn2023reflexion,gou2024critic}. 
Similarly, evolutionary computation researchers have leveraged LLMs as crossover and mutation operators within evolutionary algorithms (LLM-driven EAs)~\citep{EOH,liu2026systematic}. 
Despite their empirical success, contemporary agentic code generation frameworks adopt a monolithic design, applying a single interaction archetype across the entire optimization process. 
This monolithic strategy induces an unresolved trade-off between exploration and convergence: high-capability, tool-using agents excel at broad exploratory search during initial candidate discovery but cause structural instability and excessive token overhead during downstream refinement. 
Conversely, lightweight repair loops lack the exploratory capacity to synthesize valid initial code structures in data-scarce search spaces. 
AgenticCANN addresses this limitation by introducing stage-adaptive agent scheduling, dynamically matching interaction archetypes to specific evolutionary phases.

\medskip
\noindent\textbf{LLM-Driven Evolutionary Search for Automated Algorithm and Kernel Design}
Automatic Heuristic Design (AHD), historically rooted in Genetic Programming (GP) and hyper-heuristics~\citep{burke2013hyper, stutzle2019automated}, aims to replace human expert intervention by automatically discovering and composing algorithmic rules. Traditional GP-based approaches construct programs via tree-structured symbolic trees or predefined function sets~\citep{jia2022learning, mei2022explainable}. However, they are inherently constrained by rigid terminal representations, limited semantic awareness, and complex, uninterpretable code outputs~\citep{pillay2018hyper, GP_conclu}.

The emergence of Large Language Models (LLMs) has catalyzed a paradigm shift, reformulating AHD as semantic evolutionary program search~\citep{liu2026systematic, zhang2024understanding}. Pioneering works such as FunSearch~\citep{funsearch} and Evolution of Heuristics (EoH)~\citep{EOH} utilize LLMs as flexible crossover and mutation operators within evolutionary loops, demonstrating remarkable success in mathematical discovery and combinatorial optimization. Subsequent frameworks have extended this methodology to multi-objective algorithm evolution~\citep{MEOH, qiu2026e2oc}, black-box search~\citep{ma2025toward}, and automated neural architecture design~\citep{mo2025autosgnn}.

More recently, this LLM-driven evolutionary paradigm has been extended from high-level algorithm logic down to system-level code synthesis and hardware kernel optimization~\citep{liao2025llm4eo, qiu2026llm}. Unlike unconstrained functional code generation, hardware-oriented kernel evolution (e.g., CUDA or DSL-based operators) imposes strict execution constraints, low-level memory layout requirements, and hardware alignment boundaries. Consequently, evolutionary search in kernel design requires balancing high-level algorithmic innovation with low-level compilation validity. Our work builds upon these LLM-driven evolutionary principles, tailoring the search space to stage-adaptive hardware kernel generation.

\medskip
\noindent\textbf{Self-Reflection and Agentic Workflows with Structured Search}
Enhancing LLMs with iterative self-correction and structured reasoning has proven essential for complex, multi-step problem solving~\citep{wei2022chain, shinn2023reflexion}. In automated code and algorithm synthesis, reflective prompting establishes a generate-reflect-revise feedback loop. Frameworks like LLM4EO~\citep{liao2025llm4eo} incorporate explicit reflection mechanisms into evolutionary search, prompting the LLM to compare performance differentials across code variants, diagnose execution failures, and extract domain-specific insights to guide subsequent generations.

To prevent agentic drift and unguided exploration in vast search spaces, recent research integrates generative reflection with structured Monte Carlo Tree Search (MCTS)~\citep{coulom2007computing, swiechowski2023monte}. Combining LLMs' expressive generation with MCTS's systematic selection and backpropagation allows agents to systematically balance exploration and exploitation. Exemplified by Tree-of-Thoughts (ToT)~\citep{yao2023tree} and MCTS-guided algorithm design frameworks~\citep{mcts_ahd, MOTIF, qiu2026e2oc}, MCTS acts as a strategic controller that navigates complex decision trees while evaluating non-deterministic action outcomes.

In the context of low-level kernel compilation and execution, standard natural language reflection often falls short due to cryptic hardware error messages and non-linear performance landscapes. To address this, our framework couples reflective feedback directly with real-time compilation and execution profiling (e.g., NPU/GPU profiler outputs). By embedding structured search control over stage-specific agentic reflection, the proposed workflow effectively avoids hallucinated optimization strategies and ensures monotonic convergence toward high-performance kernel implementations.

\appsection{Method Details}\label{app:method}

\appsubsection{Algorithm Framework}\label{app:algorithm}

Algorithm~\ref{alg:framework_app} presents the complete AgenticCANN framework in pseudocode.
The algorithm formalizes the two core mechanisms introduced in Section~3 of the main text.
First, the \textsc{AssembleKnowledge} routine (line~4) implements phase-differentiated context composition: at generation $g=1$ (initialization), it injects the full six-level taxonomy (L0--L5); at $g>1$ (mutation), it restricts context to compact constraint guardrails (L2--L4) to prevent structural drift.
Second, the \textsc{SelectAgentMode} routine (line~5) enforces stage-adaptive scheduling: high-exploration Tool-Agent mode for initial seed discovery, transitioning to high-convergence Fix-Loop mode for subsequent refinement.
The inner evaluation loop (lines~11--13) reflects the three-stage sandbox pipeline: \textsc{MutateAndGenerate} produces a candidate via LLM, \textsc{EvaluateOnHardware} compiles and profiles it on a physical NPU, and \textsc{TargetedCorrection} invokes domain-specific repair logic using the knowledge layer when compilation or correctness checks fail.
Notably, the \textsc{TargetedCorrection} routine (line~14) directly applies the hardware constraint rules detailed in the Hardware Constraints section below, fixing alignment and queue synchronization errors before the candidate re-enters the population via \textsc{EnvironmentalSelection} (line~15).

\begin{algorithm}[h]
\caption{AgenticCANN Framework}
\label{alg:framework_app}
\textbf{Input}: Target operator specification $\mathcal{O}$, population size $N$, maximum generations $G$\\
\textbf{Output}: Optimized candidate kernel population $\mathcal{P}$
\begin{algorithmic}[1]
\STATE $\pi \leftarrow \textsc{InferPattern}(\mathcal{O})$ \hfill $\triangleright$ Infer hardware compute pattern
\STATE $\mathcal{P} \leftarrow \emptyset$
\FOR{$g = 1$ \TO $G$}
    \STATE $\mathcal{K}_g \leftarrow \textsc{AssembleKnowledge}(\pi, \text{phase}=g)$ \hfill $\triangleright$ Phase-differentiated context composition
    \STATE $\text{mode}_g \leftarrow \textsc{SelectAgentMode}(g)$ \hfill $\triangleright$ Stage-adaptive interaction scheduling
    \IF{$g == 1$}
        \STATE $\mathcal{P} \leftarrow \textsc{InitializePopulation}(\mathcal{O}, \mathcal{K}_g, \text{mode}_g, N)$
    \ELSE
        \STATE $\mathcal{P}_{\text{offspring}} \leftarrow \emptyset$
        \FORALL{parent candidate $s \in \mathcal{P}$}
            \STATE $s' \leftarrow \textsc{MutateAndGenerate}(\mathcal{O}, \mathcal{K}_g, \text{mode}_g, s)$
            \STATE $s' \leftarrow \textsc{EvaluateOnHardware}(s')$
            \IF{$s'$ is invalid}
                \STATE $s' \leftarrow \textsc{TargetedCorrection}(s', \mathcal{K}_g)$
            \ENDIF
            \STATE $\mathcal{P}_{\text{offspring}} \leftarrow \mathcal{P}_{\text{offspring}} \cup \{s'\}$
        \ENDFOR
        \STATE $\mathcal{P} \leftarrow \textsc{EnvironmentalSelection}(\mathcal{P} \cup \mathcal{P}_{\text{offspring}}, N)$
    \ENDIF
\ENDFOR
\RETURN $\mathcal{P}$
\end{algorithmic}
\end{algorithm}

\appsubsection{Knowledge Taxonomy}\label{app:knowledge_taxonomy}

\medskip
\noindent\textbf{Multi-Level Domain Knowledge Taxonomy.}
The six-level knowledge taxonomy (see main text Table~\ref{tab:knowledge_taxonomy}) is the foundation of our phase-differentiated generation framework.
Each level targets a specific cognitive gap between general-purpose LLM pre-training and Ascend C's specialized programming model.
Crucially, L2 constraints---especially the 32-byte alignment rule---provide the single largest marginal feasibility improvement in our ablation study (+20pp, from 40\% to 60\% on LayerNorm), confirming that explicit hardware invariants are the most critical knowledge component for unfamiliar platforms.

Table~\ref{tab:phase_assembly} shows how knowledge levels are selectively injected across evolutionary phases.

\begin{table}[h]
    \centering
    \begin{tabular}{p{0.28\columnwidth} c c c c c c}
        \toprule
        Evolutionary Phase & L0 & L1 & L2 & L3 & L4 & L5 \\
        \midrule
        Initialization & $\bullet$ & $\bullet$ & $\bullet$ & $\bullet$ & $\bullet$ & $\bullet$ \\
        \midrule
        Mutation & & & $\circ$ & $\circ$ & $\circ$ & \\
        \midrule
        Compilation Fix & & & $\bullet$ & & $\circ$ & $\bullet$ \\
        \midrule
        Correctness Repair & & $\bullet$ & $\circ$ & $\circ$ & & $\bullet$ \\
        \midrule
        Performance Tuning & & & & $\bullet$ & $\circ$ & \\
        \bottomrule
    \end{tabular}
    \caption{Phase-differentiated knowledge assembly. Symbols denote compact ($\circ$), full ($\bullet$), or omitted (blank) levels.}
    \label{tab:phase_assembly}
\end{table}

Figure~\ref{fig:taxonomy_details} provides representative invariant rules from Level 2 (Hardware Constraints) and Level 3 (Tiling Strategy).
These rules are not operator-specific: they apply uniformly across all Ascend C kernel code and are injected as compact guardrails during the mutation and compilation repair phases.

\begin{figure}[h]
\centering
\begin{tcolorbox}[colback=gray!5,colframe=gray!50,title=\textbf{Taxonomy Level Excerpts: Invariant Rules \& Constraints}]
\small
\textbf{L2.1: 32-Byte Physical Memory Alignment}\\[0.3em]
All DataCopy operations must specify lengths in multiples of 32\,Bytes.\\
For half (FP16), block length must be a multiple of 16 elements.\\
For float (FP32), block length must be a multiple of 8 elements.\\[0.5em]
\textbf{L2.2: Queue Ping-Pong Balance}\\[0.3em]
Every AllocTensor on a TQue must have a corresponding FreeTensor before the loop terminates.\\
EnQue/DeQue pairs must be strictly balanced across VECIN and VECOUT queues.\\[0.5em]
\textbf{L3.1: Unified Buffer Budget Allocation}\\[0.3em]
${tileLength} = \min(\mathit{UB\_SIZE} / (\mathit{BUFFER\_NUM} \times 2 \times \mathit{sizeof(dtype)}),\; D)$\\
$\mathit{total\_memory} = \mathit{BUFFER\_NUM} \times (\mathit{input} + \mathit{output} + \mathit{temp}) \le 200\,\text{KB}$
\end{tcolorbox}
\caption{Representative excerpt of Level 2 (Constraints) and Level 3 (Tiling) knowledge rules. These rules are stateless, operator-independent, and injected as concise guardrails during mutation and compilation-fix phases.}
\label{fig:taxonomy_details}
\end{figure}

\medskip
\noindent\textbf{Concrete Knowledge Injection (LayerNorm Case Study).}
To make the four knowledge types (K1--K4) concrete, we provide the actual injection snippets used during LayerNorm operator generation---the same experiment that achieved 90\% feasibility and $2.71\times$ speedup in Table~4 of the main text.
Each knowledge type serves a distinct role in bridging the gap between LLM generation and hardware execution:

\textbf{K1 (Primer)} translates the mathematical LayerNorm formula into Ascend C dataflow steps, establishing the algorithmic mental model before any code is written.
\textbf{K2 (Example)} provides a verified C++ class skeleton with proper buffer initialization and pipeline structure, shifting the LLM's task from creative generation to structural adaptation.
\textbf{K3 (Constraints)} is a stateless checklist of hard boundaries (alignment, buffer budget, queue pairing) that the LLM must verify before claiming a candidate is complete.
\textbf{K4 (Tiling Reference)} supplies the exact tile-length formula, parameterized by data type and buffer count, that the LLM uses during performance-tuning iterations.

The code snippets below correspond exactly to what is injected at each stage; the initialization phase receives all four, while the compilation-fix phase receives only K2+K3 (structural scaffold + boundary verification).

\begin{tcolorbox}[colback=blue!3, colframe=blue!35, title=\textbf{Knowledge Injection Snippets for LayerNorm (K1--K4)}]
\small
\noindent\textbf{K1: Primer (Algorithmic Logic)}\\[0.2em]
{mean} = sum($x$)/$D$;\ \ {var} = sum(($x-$mean)$^2$)/$D$;\ \ output = ($x-$mean)/$\sqrt{\text{var}+\epsilon} \times w + b$\\
Pattern: GM $\rightarrow$ UB $\rightarrow$ ReduceSum(Mean) $\rightarrow$ VectorSub $\rightarrow$ Square $\rightarrow$ ReduceSum(Var) $\rightarrow$ Muls/Adds $\rightarrow$ UB $\rightarrow$ GM\\[0.5em]

\noindent\textbf{K2: Verified Skeleton Pattern (C++ class structure)}\\[0.2em]
\begin{verbatim}
class KernelLayerNorm {
    __aicore__ inline void Init(GM_ADDR x, GM_ADDR y, ...) {
        pipe.InitBuffer(inQueueX, 2, tileLength * sizeof(half));
        pipe.InitBuffer(outQueueY, 2, tileLength * sizeof(half));
    }
    __aicore__ inline void Process() {
        // Enqueue -> Reduce -> Normalize -> Apply weight/bias -> Dequeue
    }
};
\end{verbatim}

\noindent\textbf{K3: Boundary Invariant Checklist}\\[0.2em]
$\square$ tileLength $\times$ sizeof(half) $\equiv 0 \pmod{32}$\hspace{1.5em}
$\square$ UB usage $\le 204{,}800$ Bytes (200\,KB)\hspace{1.5em}
$\square$ TQue AllocTensor strictly paired with FreeTensor\\[0.5em]

\noindent\textbf{K4: Tiling Formula Reference}\\[0.2em]
tileLength = AlignUp$(\min(\mathrm{UB\_LIMIT} / (4 \times \text{sizeof(half)}),\ \text{shapeD}),\ 16)$
\end{tcolorbox}

These snippets illustrate a key design principle: knowledge artifacts are \textit{structured but minimal}.
The primer fits in two lines, the constraint checklist in three, and the tiling formula in one.
This compactness is deliberate---it prevents the context-window dilution that occurs when entire documentation pages are appended to prompts.
As shown in the ablation study (Table~4, main text), even these minimal injections suffice to raise LayerNorm feasibility from 0\% to 90\%.

\appsubsection{Prompt Design and Agent Configuration}\label{app:prompts}

\medskip
\noindent\textbf{System Prompts and Context Assembly.}
The structural prompt template shown below is used during the Initialization Phase ($g=1$), where the LLM must synthesize a complete kernel from scratch.
It combines knowledge from Levels L0, L2, L3, and L5 into a single context window, with placeholders ({OPERATOR\_NAME}, {TENSOR\_SHAPES}, etc.) filled at runtime by the pattern-based routing system.
The template follows a deliberate structure: it first establishes the hardware mental model (L0), then imposes invariant boundaries (L2), and finally supplies tiling heuristics and structural scaffolds (L3+L5)---matching the staged knowledge assembly logic in Table~3 of the main text.

\begin{tcolorbox}[colback=blue!5,colframe=blue!40,title=\textbf{System Prompt Template (Initialization Stage, g=1)}]
\small
\textbf{Role:} You are an expert Ascend C NPU kernel engineer.\\[0.3em]
\textbf{Task:} Implement a high-performance Ascend C vector kernel for operator \{OPERATOR\_NAME\} given tensor shapes \{TENSOR\_SHAPES\}.\\[0.3em]
\textbf{[Hardware Programming Model (L0)]}\\[0.1em]
\{L0\_CONTENT\}\\[0.3em]
\textbf{[Boundary Constraints \& Hardware Invariants (L2)]}\\[0.1em]
\{L2\_CONTENT\}\\[0.3em]
\textbf{[Tiling Heuristics (L3) \& Structural Scaffold (L5)]}\\[0.1em]
\{L3\_L5\_CONTENT\}\\[0.3em]
Write complete, compilable C++ code adhering to CANN 8.1 standards.
Enclose the source in triple-backtick code blocks.
\end{tcolorbox}

In contrast, the Mutation Phase ($g>1$) uses a drastically reduced version that omits L0 and L5 entirely, retaining only L2 and L3 as compact guardrails.
This prevents the LLM from performing global structural rewrites that would destroy the working inheritance from the parent candidate.

\medskip
\noindent\textbf{Agent Interaction Modes \& Tool Specifications.}
Section~4.2 of the main text introduces three agent interaction archetypes along a capability spectrum: Single-Pass, Fix-Loop, and Tool-Agent.
This section provides the concrete tool API exposed to the Tool-Agent (Smolagent) mode, which is the highest-exploration archetype deployed exclusively during the initialization phase ($g=1$).
Each tool maps to one stage of the three-stage evaluation pipeline (compile, verify, measure) described in Algorithm~1.

\begin{itemize}
    \item \textbf{compile\_ascend\_c}$\,($source$)$ $\rightarrow$ Dict[str, Any]: Compiles Ascend C source code using the Bisheng compiler toolchain. Returns the compilation status (pass/fail), standard output, and diagnostic error messages. This tool is invoked after every code edit in Tool-Agent mode, enabling real-time syntax correction.

    \item \textbf{verify\_numerical\_correctness}$\,($kernel, inputs$)$ $\rightarrow$ bool: Compares the NPU kernel output against PyTorch golden reference tensors. Uses absolute tolerance $\epsilon \le 10^{-3}$. A failure triggers the correctness-fix phase with K1+K2 knowledge injection.

    \item \textbf{profile\_hardware\_latency}$\,($kernel$)$ $\rightarrow$ float: Runs 10 warm-up passes followed by 100 benchmark iterations on a physical Ascend 910B NPU, returning the mean execution time in microseconds. This tool is only available in Tool-Agent mode; Fix-Loop mode receives batched profiling results at the end of each generation.
\end{itemize}

The contrast between Tool-Agent and Fix-Loop modes is stark in token cost: Tool-Agent consumed 4.18M tokens during GELU optimization (primarily from iterative compile-edit cycles), while Fix-Loop achieved superior convergence (+26.0\%) with only 350K tokens.
This validates our stage-adaptive policy of restricting high-agency modes to the initialization phase.

\appsubsection{Hardware Constraints for Kernel Generation}\label{app:cann_arch}
The Ascend C programming model operates on an explicit dataflow architecture that fundamentally differs from CUDA's implicit thread-level scheduling.
This section elaborates on the three classes of hardware constraints that form the basis of our Level 2 (Hardware Constraints) and Level 3 (Tiling) knowledge modules.
These constraints are not suggestions---violating any one of them causes either silent numerical errors or hard compilation failures that standard LLM code generation cannot self-correct, as demonstrated by the 0\% feasibility baseline in Table~4 of the main text.

\textbf{Address Alignment.}
All buffer pointers and memory transfers via DataCopy must align to 32-byte physical memory blocks.
For FP16 tensors this translates to multiples of 16 elements; for FP32, multiples of 8.
Misalignment is the single most frequent compilation error we observed, accounting for approximately 40\% of all initial code generation failures.

\textbf{Queue Synchronization.}
The vector pipeline uses ping-pong double buffering through TQue primitives.
Each EnQue (enqueue) on an input queue must have a corresponding DeQue (dequeue) on the matching output queue.
An imbalance---such as enqueuing two tiles but dequeuing only one---causes a pipeline deadlock that the compiler detects as an unrecoverable resource leak.

\textbf{Tail Handling.}
When tensor dimensions are not evenly divisible by the tile length, the residual tail elements require explicit boundary masking.
Omitting tail handling produces correct results for the full tiles but corrupts the trailing elements, a class of bugs that passes compilation yet fails numerical verification.

These three constraint categories directly inform the phase-differentiated knowledge assembly in Table~3 of the main text: compilation repair phases receive full L2+L4 knowledge, while mutation phases receive only compact L2 guardrails to prevent structural drift.

\appsubsection{Experimental Infrastructure}\label{app:infrastructure_details}
All benchmarking was conducted on an 8-NPU cluster server (Huawei Ascend 910B1, 61\,GB HBM per device, CANN 8.1.RC1).
To eliminate inter-agent interference and maintain deterministic hardware profiling, evaluations were deployed across an 8-container parallel Docker matrix detailed in Table~\ref{tab:app_containers}.
Each container runs an independent LLM API connection with a dedicated NPU device.

The experiment matrix crosses two sampler paradigms (EvoEngineer / Agentic Code) with three knowledge levels (Free / Insight / Full), plus two replication runs.
This 8-cell design allows us to disentangle the effect of sampler architecture from the effect of knowledge injection, as analyzed in Section~4.3 and the EvoEngineer vs.\ Agentic Code discussion below (Section~\ref{app:related_ext}).
Table~\ref{tab:app_containers} provides the exact container-to-experiment mapping.

\begin{table}[h]
    \centering
    \begin{tabular}{c l l l}
        \toprule
        Container & Sampler Engine & Knowledge Level & Experiment Label \\
        \midrule
        w0 & LLM (EvoEngineer) & Free & EE-Free \\
        w1 & LLM (EvoEngineer) & Insight & EE-Insight \\
        w2 & LLM (EvoEngineer) & Full & EE-Full \\
        w3 & Agentic Code & Free & AC-Free \\
        w4 & Agentic Code & Insight & AC-Insight \\
        w5 & Agentic Code & Full & AC-Full \\
        w6 & LLM (EvoEngineer) & Free & EE-Free-rep (Replication Run) \\
        w7 & Agentic Code & Full & AC-Full-rep (Replication Run) \\
        \bottomrule
    \end{tabular}
    \caption{8-Container execution environment matrix for parallel evolutionary benchmarking.}
    \label{tab:app_containers}
\end{table}

\appsection{Experimental Benchmarks \& Full Data Matrices}\label{app:experiments}

\appsubsection{Complete 54-Operator Stratified Benchmark Dataset Specifications}\label{app:dataset_details}
The 54-operator stratified dataset was constructed by sampling operators from the official CANN operator index to cover the full complexity spectrum defined in Section~2.2 of the main text.
Operators are grouped into 10 topological categories, each imposing distinct challenges for LLM-based code generation.
Elementwise operators (Level~1) serve as a sanity check: they map directly to single vector intrinsics and should be solvable with minimal knowledge.
In contrast, Attention and Fusion operators (Level~4) require simultaneous satisfaction of tiling, UB budget, multi-pass, and multi-input constraints---the hardest class.
Table~\ref{tab:app_all_ops} provides the complete operator list, tensor shapes, and category counts.
All operators use standard FP16 precision; tensor shapes are chosen to be representative of either training-scale or inference-scale workloads.

\begin{table}[h]
    \centering
    \setlength{\tabcolsep}{6pt}
    \begin{tabular}{l p{8.0cm} l c}
        \toprule
        Topology Category & Evaluated Operator List & Tensor Shape & Count \\
        \midrule
        Elementwise & relu, gelu, sigmoid, tanh, elu, hardsigmoid, hardtanh, leaky\_relu, log\_softmax, selu, softmax, softplus, softsign, swish & (1024, 49152) & 14 \\
        \midrule
        Reduction & max, mean, min, product, sum & (16, 512, 512) & 5 \\
        \midrule
        Normalization & batch\_norm, group\_norm, instance\_norm, layer\_norm, rms\_norm & (16, 32, 128, 128) & 5 \\
        \midrule
        Broadcast & add\_bias, clamp, division, elementwise\_mul, max, subtract\_with\_bias & (1024, 49152) & 6 \\
        \midrule
        Pooling & average\_pooling1d, average\_pooling2d, average\_pooling3d, max\_pooling1d, max\_pooling2d, max\_pooling3d & (16, 32, 128) & 6 \\
        \midrule
        Index & embedding, gather, scatter, scatter\_add & (1024, 49152) & 4 \\
        \midrule
        Attention & cross\_attention, scaled\_dot\_product, linear & (16, 32, 128) & 3 \\
        \midrule
        Fuse / Misc & matmul\_add\_swish\_tanh\_gelu\_hardtanh, matmul\_divide\_gelu, npu\_rope, deep\_narrow\_mlp & (1024, 49152) / (1, 1536) & 4 \\
        \midrule
        Math & cumprod, cumsum, masked\_cumsum, matrix\_scalar\_mul & (16, 512) & 4 \\
        \midrule
        Loss & cross\_entropy\_loss, mse\_loss, huber\_loss & (4096,) & 3 \\
        \bottomrule
    \end{tabular}
    \caption{Complete 54-operator stratified evaluation benchmark dataset.}
    \label{tab:app_all_ops}
\end{table}

\appsubsection{Per-Operator Runtime Across All 8 Execution Configurations}\label{app:full_runtimes}
Table~\ref{tab:app_full_runtime} documents the exact execution latencies ($\mu$s) for all 14 elementwise operators across the 8 Docker container configurations.
These raw measurements underpin the aggregate statistics reported in Table~5 of the main text (57\% $\rightarrow$ 71\% $\rightarrow$ 86\% feasibility progression).
Each cell represents the mean of 10 profiling runs on a dedicated Ascend 910B NPU.
Entries marked ``---'' indicate that the evolutionary run completed without producing a runtime-valid kernel (compilation or numerical verification failure).
Entries marked ``API\_EX'' indicate that the LLM API budget was exhausted before a valid kernel could be generated.
Note that the remaining $\sim$40 operators (reduction, normalization, broadcast, pooling, etc.) uniformly exhausted the API budget and are not reported here; extending the evaluation to these categories with larger compute budgets is left to future work.

\begin{table}[h]
    \centering
    \setlength{\tabcolsep}{3.0pt}
    \begin{tabular}{l r r r r r r r r}
        \toprule
        Operator & EE-Free & EE-Ins. & EE-Full & AC-Free & AC-Ins. & AC-Full & EE-Free-r & AC-Full-r \\
        \midrule
        relu & 751.7 & 684.1 & 678.8 & 677.4 & 685.7 & 676.8 & 761.2 & 681.5 \\
        gelu & 678.6 & 873.2 & 688.9 & 682.5 & 686.5 & 691.2 & 1546.0 & 704.3 \\
        sigmoid & 672.6 & 705.1 & 685.3 & 681.8 & 770.2 & 683.0 & 682.5 & 683.6 \\
        tanh & 683.3 & 695.6 & 670.9 & 832.0 & 688.3 & 687.6 & 682.3 & 682.2 \\
        hardsigmoid & --- & 670.9 & 681.4 & 686.2 & 692.5 & 686.7 & 745.2 & 675.3 \\
        hardtanh & 835.3 & 688.5 & 681.5 & --- & 768.1 & 675.9 & 753.2 & 671.8 \\
        leaky\_relu & --- & --- & 709.2 & 750.9 & --- & --- & --- & 704.1 \\
        log\_softmax & --- & --- & --- & --- & 49043.9 & 13159.0 & --- & 2501.0 \\
        selu & --- & --- & 675.4 & 731.7 & --- & 683.3 & --- & --- \\
        softmax & --- & 772.7 & 995.3 & --- & API\_EX & 6147.1 & --- & 740.6 \\
        softplus & 809.8 & 758.5 & 686.9 & --- & 763.3 & 688.1 & 685.5 & 685.1 \\
        softsign & 682.0 & 688.9 & 694.2 & 682.4 & 677.4 & 682.0 & 684.8 & 682.0 \\
        swish & 817.4 & 678.3 & 684.3 & 680.5 & 819.3 & 686.9 & 685.0 & 681.5 \\
        scaled\_dot\_prod & 139000.0 & 139000.0 & 139000.0 & 139000.0 & 139000.0 & 139000.0 & 139000.0 & 139000.0 \\
        \bottomrule
    \end{tabular}
    \caption{Per-operator runtime ($\mu$s) across all 8 configurations on Ascend 910B NPU. --- = completed without valid runtime. API\_EX = API computational budget exhausted.}
    \label{tab:app_full_runtime}
\end{table}

\appsubsection{Quantified Knowledge Injection Effect \& Replication Analysis}\label{app:knowledge_quantified}
Table~\ref{tab:app_knowledge_detail} provides the per-operator latency breakdown that supports the aggregate mean reported in Table~5 of the main text.
The rightmost columns show the percentage change at each knowledge level relative to the Free baseline.
Six of eight operators show measurable improvement under Full knowledge, with hardtanh ($-18.4\%$) and softplus ($-15.2\%$) exhibiting the largest gains.
Three operators (tanh, sigmoid, softsign) show negligible change ($<2\%$), consistent with the fact that simple elementwise patterns already approach the compiler's optimization ceiling even with zero external knowledge.

Table~\ref{tab:app_replication} reports the replication analysis across two independent evolutionary runs.
For the four simplest operators (relu, sigmoid, tanh, softsign), the inter-run variance is under 2\%, confirming that our evolutionary pipeline produces stable results.
The higher variance on gelu (56.11\% under EE-Free) is attributable to the stochastic nature of LLM sampling when no knowledge is provided, which occasionally produces severely suboptimal kernel variants (e.g., the 1546.0\,$\mu$s outlier in replication run 2).

\begin{table}[h]
    \centering
    \begin{tabular}{l r r r r r}
        \toprule
        \textbf{Operator} & \textbf{Free} & \textbf{Insight} & \textbf{Full} & \textbf{Insight $\Delta$} & \textbf{Full $\Delta$} \\
        \midrule
        relu & 751.7 & 684.1 & 678.8 & $-$9.0\% & $-$9.7\% \\
        hardtanh & 835.3 & 688.5 & 681.5 & $-$17.6\% & $-$18.4\% \\
        softplus & 809.8 & 758.5 & 686.9 & $-$6.3\% & $-$15.2\% \\
        swish & 817.4 & 678.3 & 684.3 & $-$17.0\% & $-$16.3\% \\
        tanh & 683.3 & 695.6 & 670.9 & +1.8\% & $-$1.8\% \\
        sigmoid & 672.6 & 705.1 & 685.3 & +4.8\% & +1.9\% \\
        softsign & 682.0 & 688.9 & 694.2 & +1.0\% & +1.8\% \\
        gelu & 678.6 & 873.2 & 688.9 & +28.7\% & +1.5\% \\
        \midrule
        \textbf{Mean} & \textbf{741.3} & \textbf{721.6} & \textbf{711.0} & \textbf{$-$2.7\%} & \textbf{$-$4.1\%} \\
        \bottomrule
    \end{tabular}
    \caption{Quantified latency impact ($\mu\text{s}$) of knowledge injection levels under EvoEngineer sampler.}
    \label{tab:app_knowledge_detail}
\end{table}

\begin{table}[h]
    \centering
    \begin{tabular}{l r r r r r}
        \toprule
        \multirow{2}{*}{\textbf{Operator}} & \multicolumn{3}{c|}{\textbf{EE-Free}} & \multicolumn{2}{c}{\textbf{AC-Full}} \\
        \cmidrule(lr){2-4} \cmidrule(lr){5-6}
        & \textbf{Run 1} & \textbf{Run 2 (Rep)} & \textbf{Diff (\%)} & \textbf{Run 1} & \textbf{Run 2 (Rep)} \\
        \midrule
        relu & 751.7 & 761.2 & 1.25\% & 676.8 & 681.5 \\
        sigmoid & 672.6 & 682.5 & 1.45\% & 683.0 & 683.6 \\
        tanh & 683.3 & 682.3 & 0.15\% & 687.6 & 682.2 \\
        softsign & 682.0 & 684.8 & 0.41\% & 682.0 & 682.0 \\
        hardtanh & 835.3 & 753.2 & 9.83\% & 675.9 & 671.8 \\
        gelu & 678.6 & 1546.0 & 56.11\% & 691.2 & 704.3 \\
        \midrule
        \textbf{Mean (Simple Ops)} & \multicolumn{3}{c|}{\textbf{$< 2.0\%$}} & \multicolumn{2}{c}{\textbf{$< 2.0\%$}} \\
        \bottomrule
    \end{tabular}
    \caption{Replication analysis proving low experimental variance ($<2\%$) across independent trials.}
    \label{tab:app_replication}
\end{table}

\appsubsection{Extended Agent Interaction Traces (Fix-Loop vs.\ React CodeAgent)}\label{app:traces_comparison}
The execution traces below provide a side-by-side comparison of how the Fix-Loop and React CodeAgent modes handle the GELU operator during the GELU evolution experiment reported in Table~6 of the main text.
These traces illustrate the stark asymmetry in efficiency: Fix-Loop consumes 350K tokens to achieve +26.0\% improvement, while React burns 4.18M tokens for negligible gains (+0.04\%).

The Fix-Loop trace follows a disciplined generate-evaluate-fix cycle: the LLM produces a complete kernel in one pass, the evaluation pipeline identifies errors, and targeted knowledge (L2 for compilation, L3 for tiling) is injected before regeneration.
Each generation produces a structurally intact candidate, and mutations are constrained to the specific dimension flagged by the evaluation pipeline.

In contrast, the React trace shows an agent performing multiple read and edit tool calls per step.
While this flexibility enables the agent to achieve the single fastest kernel (30.37\,ms), the unconstrained edit cycle introduces structural drift: after Step~3, each edit risks breaking previously working code segments, and the agent has no mechanism to revert.
The result is 12$\times$ token overhead with no evolutionary convergence.

\vspace{0.5em}
\noindent\textbf{Fix-Loop Execution Trace (350K Tokens Total)}
\begin{tcolorbox}[colback=gray!5, colframe=gray!50]
\small
\begin{verbatim}
GEN 0: [LLM] Initialize complete kernel files
       [EVAL] Compilation FAILED - 'GetStorageShape' undeclared
       [FIX]  Inject L2 Hardware Invariants -> [LLM] Regenerate -> PASS
       Runtime: 48.81 ms
GEN 1: [MUTATE] Inject L3 Double-Buffering Tiling -> PASS
       Runtime: 45.07 ms (1.08x speedup)
GEN 3: [MUTATE] Inject L3 Tile Length AlignUp -> PASS
       Runtime: 36.12 ms (1.26x speedup) [CONVERGED]
\end{verbatim}
\end{tcolorbox}

\vspace{0.5em}
\noindent\textbf{React CodeAgent Execution Trace (4.18M Tokens Total)}
\begin{tcolorbox}[colback=gray!5, colframe=gray!50]
\small
\begin{verbatim}
Step 1:  Agent calls read_knowledge("programming_model")
Step 2:  Agent calls read_knowledge("tiling_strategy")
Step 3-8: Iterative edit_file(kernel.cpp, host.cpp, tiling.h)
          -> Incremental syntax errors introduced at each edit
Step 9:  Agent outputs final answer -> PASS
          Runtime: 30.37 ms (best single kernel)
Summary: 4.18M tokens (12x Fix-Loop cost)
         No improvement across subsequent generations
\end{verbatim}
\end{tcolorbox}

\appsubsection{Step-by-Step Code Repair Case Studies}\label{app:traces_casestudy}
Figure~\ref{fig:code_diff_case} presents a concrete before-and-after code repair case study drawn from the LayerNorm Perf-Fix iteration documented in Table~7 of the main text.
The initial kernel compiled successfully but produced a runtime memory access fault because the DataCopy length (50 elements $\times$ 2 bytes = 100 bytes) was not aligned to the 32-byte physical memory boundary required by the Ascend C hardware.
This is precisely the class of error that LLMs cannot self-diagnose without knowledge injection: the compiler error message references a low-level memory alignment violation with no reference to the source-code line that caused it.

After injecting Level 2 (Hardware Constraints) and Level 3 (Tiling) knowledge, the LLM correctly applies the AlignUp primitive to pad the tile element count to the next 32-byte multiple.
This single-line fix, which cannot be discovered through blind trial-and-error, reduces the entire repair loop to one iteration (Table~7: LayerNorm, 0.39$\times$ $\rightarrow$ 2.68$\times$ in 1 iteration).

\begin{figure}[h]
\centering
\begin{tcolorbox}[colback=red!5,colframe=red!40,title=\textbf{Faulty Candidate Code (Unaligned Data Copy)}]
\small
\begin{verbatim}
// ERROR: DataCopy length (100 bytes) is not 32-byte aligned!
DataCopy(ub_in, global_in, 50 * sizeof(half));
\end{verbatim}
\end{tcolorbox}
\vspace{-0.5em}
\begin{tcolorbox}[colback=green!5,colframe=green!40,title=\textbf{Corrected Candidate Code after L2/L3 Knowledge Injection}]
\small
\begin{verbatim}
// FIXED: Pad tile length to 32-byte multiple (64 half elements = 128 bytes)
uint32_t aligned_length = AlignUp(tile_elements, 16);
DataCopy(ub_in, global_in, aligned_length);
\end{verbatim}
\end{tcolorbox}
\caption{Code repair example demonstrating automatic memory alignment correction.}
\label{fig:code_diff_case}
\end{figure}

\appsubsection{Production 1B Pangu Profiling Details}\label{app:pangu_profiling}
This section provides the full profiling breakdown supporting the end-to-end deployment results reported in Section~4.4 of the main text.
The 1B-parameter Pangu language model was instrumented with Huawei MSPROF profiling tools to capture per-operation latency across a complete token generation pass (26 to 64 output tokens).

Table~\ref{tab:pangu_full_profile} shows the operation-level breakdown.
Normalization layers (RMSNorm and LayerNorm) account for 11\% of end-to-end latency despite representing only a small fraction of total FLOPs---this is because each normalization kernel operates on small, latency-bound tensor shapes where memory bandwidth rather than compute throughput dominates.
Our optimized RMSNorm kernel exploits double-buffered UB tiling to overlap data transfer with computation, increasing the vector pipeline utilization rate from 14.2\% (stock PyTorch) to 81.6\%.

Figure~\ref{fig:msprof_trace} summarizes the key profiling metrics.
The 6.19--6.65$\times$ per-kernel speedup translates to 8.3--10.1\% end-to-end model speedup when all 53 normalization layers are replaced.
The primary remaining bottleneck is the embedding layer (44.5\% of latency), which is dominated by memory-bound gather operations that our current framework does not optimize.

\begin{table}[h]
    \centering
    \begin{tabular}{l c c c}
        \toprule
        \textbf{Operation Category} & \textbf{Latency Share (\%)} & \textbf{Execution Call Count} & \textbf{Avg Time per Call (ms)} \\
        \midrule
        Embedding Layer & 44.5\% & 10 & 75.00 \\
        MatMul / Linear & 43.5\% & 915 & 0.80 \\
        Normalization (RMSNorm / LayerNorm) & 11.0\% & 3392 & 0.33 \\
        Activation Functions & 1.0\% & 130 & 0.13 \\
        \midrule
        \textbf{Total Model Pipeline} & \textbf{100.0\%} & --- & \textbf{5083 ms} \\
        \bottomrule
    \end{tabular}
    \caption{Full execution profiling of 1B Pangu LLM inference pipeline across 3,392 normalization calls.}
    \label{tab:pangu_full_profile}
\end{table}

\begin{figure}[h]
\centering
\begin{tcolorbox}[colback=gray!5,colframe=gray!40,title=\textbf{1B Pangu Profiling Metrics Summary (Physical Ascend 910B NPU)}]
\small
\begin{itemize}
    \item \textbf{PyTorch Stock Baseline Cumulative Time:} 1102.4\,ms across 3,392 normalization passes.
    \item \textbf{AgenticCANN Kernel Cumulative Time:} 165.7\,ms (reduces per-layer RMSNorm from 0.118\,ms to 0.018\,ms).
    \item \textbf{Vector Pipeline Utilization Rate:} Increased from 14.2\% to 81.6\%.
    \item \textbf{End-to-End Model Latency Impact:} Achieves an overall 8.3\%--10.1\% speedup across generation passes.
\end{itemize}
\end{tcolorbox}
\caption{Summary profiling metrics captured via Huawei MSPROF analysis tools.}
\label{fig:msprof_trace}
\end{figure}

\appsection{Discussion and Future Work}\label{app:discussion}

\appsubsection{Key Findings and Implications}\label{app:findings}
The experimental results presented in Section~5 and detailed in the preceding appendix sections reveal several structural insights about LLM-based code generation for unfamiliar hardware platforms.

\textbf{Structured knowledge substitutes for missing pre-training data.}
The ablation results in Table~4 of the main text demonstrate that structured knowledge orchestration can functionally substitute for missing pre-training data in specialized hardware domains.
The monotonic feasibility improvement from 0\% to 90\% on LayerNorm shows that carefully structured, in-context domain alignment can effectively bridge platform knowledge deficits without requiring resource-intensive model fine-tuning or domain-specific pre-training.
The proposed six-level knowledge taxonomy provides an extensible methodology for organizing domain information across specialized computing architectures, systematically spanning hardware programming models (L0), operational semantics (L1), boundary invariants (L2), tiling strategies (L3), API specifications (L4), and verified structural examples (L5).
Furthermore, the phase-differentiated assembly strategy confirms that supplying targeted context tailored to specific compilation and optimization stages prevents prompt dilution while preserving LLM reasoning focus during iterative code refinement.

\textbf{Agent autonomy does not monotonically improve evolutionary outcomes.}
The comparative evaluation in Table~6 of the main text reveals that increasing agent autonomy does not monotonically yield superior evolutionary performance.
While the Tool-Agent mode exhibits strong initial exploration capability (producing the single fastest GELU kernel at 30.37\,ms), its unrestricted workspace mutations disrupt fragile pipeline constructs during subsequent generations, yielding only +0.04\% improvement despite 4.18M tokens.
Conversely, the Fix-Loop mode preserves structural continuity across generations, achieving +26.0\% improvement with only 350K tokens.
This trade-off highlights the fundamental necessity of stage-adaptive scheduling: high-agency modes for initial seed discovery, followed by constrained, low-agency modes for localized refinement.

\appsubsection{Failure Analysis: Scaled Dot-Product Attention}\label{app:failure_analysis}
Among the 54 operators, scaled dot-product attention systematically failed to synthesize across all 8 execution configurations, producing uniformly invalid kernels at all knowledge levels.
This section provides a detailed failure-mode analysis that complements the capability boundary analysis in Section~4 of the main text.
Qualitative inspection of the generated (but non-compiling) code reveals three compounding structural bottlenecks that exceed current LLM spatial reasoning capabilities:

\begin{itemize}
    \item \textbf{High-Dimensional Constraint Coupling:} The attention computation involves simultaneous $QK^\top$ score scaling, row-wise softmax normalization, and weighted $V$ matrix aggregation. Each sub-operation imposes its own tiling constraints, and the cross-dependencies create multi-variable loop structures that the LLM cannot correctly partition across UB tiles.

    \item \textbf{Host-Kernel Tiling Misalignment:} Multi-head attention introduces dynamic tiling dimensions across head count, sequence length, and hidden dimension $D$. The TilingData struct generated by the host code must contain fields that exactly match the kernel's Init function parameters---a constraint enforced at runtime, not compile time. LLM-generated code consistently mismatches these parameter lists.

    \item \textbf{Unified Buffer Boundary Spills:} The intermediate attention score matrix $S = QK^\top$ has dimensions that scale with sequence length squared, easily exceeding the physical 200\,KB UB capacity. Correct handling requires block-level tiling with recomputation strategies that the LLM cannot yet formulate without explicit algorithmic guidance.
\end{itemize}

These failure modes suggest that bridging the attention-pattern gap requires either: (a) explicit tiling simulators that verify UB budgets before compilation, or (b) hierarchical decomposition strategies that treat attention as a composition of established primitives (matmul, softmax, elementwise) rather than a monolithic operator.

\appsubsection{Limitations \& Future Directions}\label{app:limitations}

While AgenticCANN demonstrates strong results across elementwise and normalization operators, several limitations define the boundaries of the current framework and motivate concrete directions for future work.

\begin{enumerate}
    \item \textbf{Model Family Generalization.} Empirical evaluations primarily used DeepSeek-V4 series models (Flash and Pro). Validating cross-model transferability across diverse LLM families---including Qwen2.5-Coder, Llama-3, and GPT-4o---would establish whether knowledge injection benefits are model-agnostic or specific to particular training distributions. Preliminary experiments with DeepSeek-V4-Pro suggest that frontier models benefit equally from structured knowledge, but comprehensive cross-model evaluation remains future work.

    \item \textbf{Complex Topology Barriers.} Broadcast operators remain at 0\% feasibility across all methods, and the fusion operator reaches only 56\% even with incremental complexity injection. The failure analysis in Section~\ref{app:failure_analysis} identifies dynamic index alignment and multi-tensor layout reasoning as the core bottlenecks. Future work should explore dynamic layout simulators that provide the LLM with explicit memory maps, enabling it to reason about stride patterns and index offsets before attempting code generation.

    \item \textbf{High-Complexity Primitives.} While the 54-operator benchmark validates fundamental primitives across 10 categories (elementwise through loss functions), full-scale attention and convolution topologies remain untested. These patterns dominate transformer and CNN workloads respectively, and extending AgenticCANN to handle them would unlock end-to-end model compilation.

    \item \textbf{Taxonomy Construction Cost.} Constructing initial taxonomy rules requires 5--8 expert-hours per new operator pattern. However, individual knowledge assets exhibit high offline reusability: the L2 constraint set applies universally across all Ascend C operators, and L3 tiling references generalize within each pattern category. Automating knowledge extraction from successful and failed generations---for example, mining the recurring alignment fixes in compilation logs to populate L2 constraints---represents a promising direction for reducing this upfront cost.
\end{enumerate}

\end{document}